\RequirePackage{fix-cm}
\documentclass[smallcondensed]{svjour3}     
\smartqed  
\usepackage{graphicx}
\usepackage{multirow}
\usepackage{algorithm,algorithmic}

\usepackage{amsmath,amssymb,amsthm,mathrsfs,amsfonts,dsfont} 
\usepackage{amsfonts}
\usepackage{float}

\newcommand{\url}[1]{\texttt{#1}}
\theoremstyle{definition}
\newtheorem{exmp}{Example}

\newtheorem{observation}{Hypothesis}
\journalname{Journal of Automated Reasoning}

\newcommand{\MathWord}[1]{\mbox{\it #1}}

\begin{document}

\title{On the Importance of Domain Model Configuration for Automated Planning Engines}

\titlerunning{Domain Model Configuration in AI Planning}        

\author{Mauro Vallati        \and Luk\'{a}\v{s} Chrpa  \and Thomas Leo McCluskey \and Frank Hutter
}


\institute{M. Vallati \at
              School of Computing and Engineering \\
              University of Huddersield, UK\\
              \email{m.vallati@hud.ac.uk}           
           \and
           L. Chrpa \at
             Artificial Intelligence Center, Czech Technical University in Prague,\\ Prague, Czech Republic \\
              Faculty of Mathematics and Physics, Charles University in Prague,\\ Prague, Czech Republic\\
              \email{chrpaluk@fel.cvut.cz}
              \and 
             T. L. McCluskey \at 
                 School of Computing and Engineering \\
              University of Huddersield, UK\\
              \email{t.l.mccluskey@hud.ac.uk} 
              \and
              F. Hutter \at
              University of Freiburg \\
       Freiburg, Germany\\
       \email{fh@cs.uni-freiburg.de}
}

\date{}

\maketitle

\begin{abstract}
The development of domain-independent planners
within the AI Planning community is leading to
``off-the-shelf'' technology that can be used in a
wide range of applications. Moreover, it allows a
modular approach --in which planners and domain
knowledge are modules of larger software applications-- that facilitates substitutions or improvements of individual modules without changing the
rest of the system. This approach also supports the
use of reformulation and configuration techniques,
which transform how a model is represented in order to improve the efficiency of plan generation.

In this article, we investigate how the performance of domain-independent planners is affected by domain model configuration, i.e., the order in which elements are ordered in the model, particularly in the light of planner comparisons. We then introduce techniques for the online and offline configuration of domain models, and we analyse the impact of domain model configuration on other reformulation approaches, such as macros.

\end{abstract}

\section{Introduction}

Automated planning is a research discipline that addresses
the problem of generating a totally- or partially-ordered sequence of actions that transform the environment from some
initial state to a desired goal state.
In the last few decades, there has been a great deal of activity in designing planning techniques and planning engines. Since 1998, the International Planning Competition (IPC)\footnote{http://ipc.icaps-conference.org} is being organised and is increasingly attracting the attention of the AI planning community. Thanks to the IPC, a large number of advanced domain-independent planning engines is now available together with a standardised language for describing planning tasks (PDDL;~\cite{pddl}).

The development of domain-independent planners within the AI Planning community is aiming to the use of this ``off-the-shelf" technology for a wide range of applications, including space exploration \cite{ai2004mapgen}, machine tool calibration \cite{DBLP:journals/eswa/ParkinsonL15}, and urban traffic control \cite{icaps17} to mention a few.
This is despite the complexity issues inherent in plan generation, which are exacerbated by the separation of planner logic from domain
knowledge. One important class of application is where the planner is an embedded component within a larger framework, with the domain model and planning problem being generated by hand, or potentially automatically. This application class has the advantage that planners which accept the same domain model language and deliver the same type of plans can be interchanged in a modular way, without any changes to the rest of the system.

This modular approach also supports the use of reformulation and configuration techniques which can automatically re-formulate,
re-represent or tune the domain model and/or problem description, while keeping to the same input language, in order to increase the efficiency of a planner
and increase the scope of problems solved. The idea is to make these techniques to some
degree independent of domain and planner (that is, applicable to a range of domains and planning engine technology), and use them to form a wrapper around a planner,
improving its overall performance for the domain to which it is applied.

In this paper, we investigate how the configuration of domain models, i.e. the order in which elements are listed in the model, affects the performance of state-of-the-art domain-independent planners. We investigate this area by focusing on domain models written in PDDL, and by considering as configurable the potential orderings of predicate declarations, the ordering of operators, and the ordering of predicates within preconditions and effects. 
We further investigate how the orderings of learned structures (macros), when added to the original domain model, can affect the performance of such planners. Given that such structures are generated by the system containing the planner, and not hand crafted (as is commonly the case in the construction of the basic domain model), it is of great interest to determine rules for placing the macros in the domain model as they are added automatically.  
Building on our prior work \cite{DBLP:conf/ijcai/VallatiHCM15,DBLP:conf/socs/VallatiCM17}, the large experimental analysis presented in this paper provides a collection of results that help to understand the impact of domain model configuration on domain-independent planners, and to provide valuable support to knowledge engineers. In particular, in this work we:

\begin{itemize}
\item Emphasise the importance and the impact on planner performance of the way in which PDDL elements are listed within domain models, particularly in the light of planning engines comparisons. 
\item Identify possible techniques for reducing such source of performance variation, in order to provide more stable comparisons of planning engines.
\item Discuss an approach for the automated optimisation of domain model configuration for improving the performance of a given planner
\item Introduce a heuristic approach for the online configuration of PDDL models that does not require any a-priori training phase.
\item Discuss the impact of domain model configuration on other reformulation approaches, such as macros.
\item Exploit a wide range of domain-independent planners and benchmarks, in order to study the impact of domain model configuration in a variety of settings.
\end{itemize}

In 2002, Howe and Dahlman \cite{HoweD02} investigated how experiment design decisions affect the performance of planning systems, noticing that changing the order of operators and pre-conditions in domain models can affect the performance of domain-independent planning engines. They showed that different planning engines can show different levels of sensitivity to this sort of representational changes. Our investigation expands their  work in a number of ways: 
(i) while their analysis dates back 18 years and was focused on STRIPS problems only, we consider a larger benchmark set, and a range of current planners and instances; (ii) while their analysis was limited to a fairly small number of permutations (10 per problem) we consider as modifiable the order in which operators, predicates, preconditions, and postconditions are listed in a domain model; (iii) we evaluate the impact of domain model configuration using various performance metrics, namely coverage, IPC score and penalised average runtime; (iv) we investigate the importance of domain model configuration when macros are involved;
(v) we introduce two approaches --one offline based on learning, and one online based on heuristics-- for configuring domain models in order to improve planners' performance; and (vi) we describe which changes have a stronger impact on planners' performance, thus providing help for engineering future domain models.

The main empirical finding in this article is that it is possible to configure domain models to improve the performance of domain-independent planning engines.
Also, our thorough experimental analysis provides 
valuable lessons from which domain engineers can learn how to encode effective and efficient domain models. Our results indicate that the configuration of domain models can lead to up-to-25-fold speedup, and that the correct positioning of a single macro operator can result in up-to-3-orders-of-magnitude runtime improvements. 

This article is organised as follows. First, we discuss related work (Section \ref{sec:related}), provide background on automated planning (Section \ref{sec:background}), and provide a formalisation of the considered domain model configuration problem (Section \ref{sec:formalisation}). Then, Section \ref{sec:impact} assesses the impact of domain model configurations by considering planners and benchmarks of the 2014 edition of the IPC. Section \ref{sec:configuration} focuses on methods and techniques for exploiting domain model configuration in order to maximise the performance of domain-independent planners, while Section 7 considers the case where learned macro-operators are automatically added to a domain model through a pre-processing stage. 
Finally, we give conclusions and present future avenues of research (Section \ref{sec:conclusion}).

\section{Related Work}\label{sec:related}


Significant work has been done in the area of reformulation for improving the performance of domain-independent planners. 
\textit{Macros} \cite{mo,ma,my_ker,McCPort} are one of the best known types of reformulation in classical planning; they encapsulate in a single planning operator a sequence of operators.  Macro-FF~\cite{macroff}, which learns macros through analysis of relations between static predicates, and Wizard~\cite{ma2}, which learns macros by genetic algorithms, are good examples of macro learning systems. Another reformulation technique are \textit{Entanglements}~\cite{my_sara,my_ecai}, which represent relations between planning operators and predicates, aimed at eliminating unpromising alternatives in a planning engine's search space. Inner Entanglements~\cite{my_ecai} are relations between pairs of operators and predicates which capture exclusivity of predicate achievement or requirement between the given operators. Outer Entanglements~\cite{my_sara,my_ecai} are relations between planning operators and predicates whose instances are present in the initial state or the goal. Finally, \textit{action schema splitting} \cite{DBLP:conf/aips/ArecesBD014} is a technique which transform operators with a large number of parameters into several smaller operators, in order to reduce the number of ground actions. 

Riddle, Holte, and Barley \cite{RiddleHB11} investigated how different encodings of the well-known BlocksWorld domain impact performance of planning engines.
In terms of impact of differently-shaped models on planners, Howe and Dahlman \cite{HoweD02} investigated how experiment design decisions affect the performance of planning systems, noticing that changing the order of elements in either domain models or problem specifications can affect performance. 

Considering the wider area of AI, the impact of the configuration of knowledge models has been recently investigated in Abstract Argumentation \cite{ijar17}. Results indicate that the configuration of the representation of the argumentation framework plays a significant role for performance.



\section{Classical Planning}\label{sec:background}

Classical planning is concerned with finding a (partially or totally ordered) sequence of actions transforming the static, deterministic and fully observable environment from the given initial state to a desired goal state~\cite{apl,pddl2.1}.

In the classical representation, a \textit{planning task} consists of a \textit{planning domain model} and a \textit{planning problem}, where the planning domain model describes the environment and defines planning operators while the planning problem defines concrete objects, an initial state and a set of goals. The environment is described by \textit{predicates} that are specified via a unique identifier and terms (variable symbols or constants). 
For example, the domain model in the ``logistics'' domain contains a predicate \textsf{at(?t ?p)}, where \textsf{at} is a unique identifier, and \textsf{?t} and \textsf{?p} are variable symbols, denoting that a truck \textsf{?t} is in a location \textsf{?p}. Predicates thus capture general relations between objects.

A \textbf{planning task} is a pair $\Pi = (Dom_\Pi,Prob_\Pi)$ where a \textbf{planning domain model} $Dom_\Pi = (P_\Pi,Ops_\Pi)$ is a pair consisting of a finite set of predicates $P_\Pi$ and planning operators $Ops_\Pi$, and a \textbf{planning problem} $Prob_\Pi=(Objs_\Pi,I_\Pi,G_\Pi)$ is a triple consisting of a finite set of objects $Objs_\Pi$, initial state $I_\Pi$ and goal $G_\Pi$.

Let $ats_\Pi$ be the set of all \textbf{atoms} that are formed from the predicates $P_\Pi$ by applying all possible substitution mappings from the predicates' parameters (variable symbols) to the objects from $Objs_\Pi$. In other words, an atom is an \textbf{instance} of a predicate (in this article, when we use the term instance, we mean an instance that is fully {\it ground}). A \textbf{state} is a subset of $ats_\Pi$, and the \textbf{initial state} $I_\Pi$ is a distinguished state. The \textbf{goal} $G_\Pi$ is a non-empty subset of $ats_\Pi$, and a \textbf{goal state} is any state that contains the goal $G_\Pi$.

Notice that the semantics of state reflects the full observability of the environment. That is, that for a state $s$, atoms present in $s$ are assumed to be true in $s$, while atoms not present in $s$ are assumed to be false in $s$. 

\textit{Planning operators} are ``modifiers'' of the environment. They consist of \textit{preconditions}, i.e., what must hold prior to an operator's application, and \textit{effects}, i.e., what is changed after its application. We distinguish between \textit{negative effects}, i.e., what becomes false, and \textit{positive effects}, i.e., what becomes true after an operator's application. \textit{Actions} are instances of planning operators, i.e., an operator's parameters, as well as corresponding variable symbols in its preconditions and effects, are substituted by objects (constants). Planning operators capture general types of activities that can be performed. Similarly to predicates that can be instantiated to atoms to capture given relations between concrete objects, planning operators can be instantiated to actions to capture given activities between concrete objects.


A \textbf{planning operator}
$o=(\MathWord{name}(o),\MathWord{pre}(o),\MathWord{eff}(o))$
is specified such that $\MathWord{name}(o) =
\MathWord{op\_name}(x_1, \dots, x_k)$, where $\MathWord{op\_name}$ is a
unique identifier and $x_1, \dots, x_k$ are all the variable symbols
(parameters) appearing in the operator,
$\MathWord{pre}(o)$ is a set of predicates representing its precondition, and $\MathWord{eff}(o)$ represents its effects, divided into $\MathWord{eff}^-(o)$ and $\MathWord{eff}^+(o)$ (i.e., $\MathWord{eff}(o)=\MathWord{eff}^-(o)\cup \MathWord{eff}^+(o)$) that are sets of predicates representing the operator's negative and positive effects, respectively. \textbf{Actions} are instances of planning operators that are formed by substituting objects, which are defined in a planning problem, for operators' parameters as well as for the corresponding variable symbols in operators' preconditions and effects. An action $a=(\MathWord{pre}(a),\MathWord{eff}^-(a)\cup\MathWord{eff}^+(a))$ is \textbf{applicable} in a state $s$ if and only if $\MathWord{pre}(a)\subseteq s$. Application of $a$ in $s$, if possible, results in a state $(s\setminus\MathWord{eff}^-(a))\cup \MathWord{eff}^+(a)$.

A solution of a planning task is a sequence of actions transforming the environment from the given initial state to a goal state.  

A \textbf{plan} is a sequence of actions. A plan is a \textbf{solution} of a planning task $\Pi$, a \textbf{solution plan} of $\Pi$ in other words, if and only if a consecutive application of the actions from the plan starting in the initial state of $\Pi$ results in the goal state of $\Pi$.

A \textbf{macro-operator} encapsulates a sequence of (primitive) planning
operators and can be represented as an ordinary planning operator. Assuming that operators $o_i$ and $o_j$ consist of some variable symbols that are the same (e.g. \textsf{unstack(?x ?y)} and \textsf{putdown(?x)}), a macro-operator $o_{i,j}$ is constructed by assembling planning operators $o_i$ and $o_j$ (in that order) in the following way:
\begin{itemize}
    \item $\MathWord{pre}(o_{i,j})=\MathWord{pre}(o_i)\cup (\MathWord{pre}(o_j)\setminus \MathWord{eff}^+(o_i))$
    \item $\MathWord{eff}^-(o_{i,j})=(\MathWord{eff}^-(o_i)\setminus \MathWord{eff}^+(o_j))\cup \MathWord{eff}^-(o_j)$
    \item $\MathWord{eff}^+(o_{i,j})=(\MathWord{eff}^+(o_i)\setminus \MathWord{eff}^-(o_j))\cup \MathWord{eff}^+(o_j)$
\end{itemize}
Clearly, $o_{i}$ must not delete any predicate required by $o_{j}$, otherwise corresponding instances of $o_{i}$
and $o_{j}$ cannot be applied consecutively. Longer macro-operators, i.e., those encapsulating longer sequences of
original planning operators can be constructed by this approach
iteratively.

Macro-operators can be thus understood as `short-cuts' in
the state space, which in some cases can speed up plan generation
considerably~\cite{ma,macroff}. This property can be useful since by exploiting them it is possible to reach the goals in fewer steps. However, the number of instances of macro-operators is often higher than the number of instances of the original operators, because they usually have more arguments deriving from the arguments of the operators that are encapsulated. This increases the branching factor in the search space, which can slow down the planning process and increase memory consumption. Therefore, it is important that benefits of macro-operators outweigh their drawbacks; this problem is known as the \textit{utility problem}~\cite{minton}. 

Macro-operators, since they are encoded in the same way as `normal' planning operators, can be added into domain models and such reformulated domain models can be passed to any planning engine. 

\section{Domain Model Configuration}\label{sec:formalisation}

\begin{figure}[!t]
\begin{minipage}{0.48\textwidth}
\begin{verbatim}
(define (domain blocksworld)
  (:predicates 
	       (on ?x ?y)
	       (ontable ?x)
	       (clear ?x)
	       (handempty)
	       (holding ?x)
  )

  (:action pick-up
	     :parameters (?x)
	     :precondition (and (clear ?x)
	                        (ontable ?x)
	                        (handempty)
	                   )
	     :effect (and (not (ontable ?x))
	                  (not (clear ?x))
	                  (not (handempty))
	                  (holding ?x)
	             )
  )
  (:action put-down
	....
  )
  (:action stack
	....
  )
  (:action unstack
    ....
  )
)
\end{verbatim}

\end{minipage}
\hfill
\begin{minipage}{0.48\textwidth}
\begin{verbatim}
(define (domain blocksworld)
  (:predicates 
	       (on ?x ?y)
	       (clear ?x)
	       (handempty)
	       (holding ?x)
	       (ontable ?x)
  )

  (:action unstack
    ....
  )
  (:action stack
	....
  )
  (:action pick-up
	     :parameters (?x)
	     :precondition (and (handempty)
	                        (clear ?x)
	                        (ontable ?x)
	                   )
	     :effect (and (holding ?x)
	                  (not (handempty))
	                  (not (ontable ?x))
	                  (not (clear ?x))
	             )
  )
  (:action put-down
	....
  )
)
\end{verbatim}

\end{minipage}

\caption{An example of two configured BlocksWorld domain models. Note that the operator schema is depicted only for the \texttt{pick-up} operator.}\label{fig:example}
\end{figure}


A key product of knowledge engineering for automated planning is the domain model. Currently, the knowledge engineering process is somewhat ad-hoc, where the skills of engineers may significantly influence the quality of the domain model, and therefore the performance of planners. Although usually underestimated, there is a significant number of degrees of freedom in domain model descriptions. 

Whereas the definitions of planning domain model and planning operator consider \emph{plain sets} of elements, i.e., it is unimportant how the elements are ordered, their order might have a significant impact on planners' performance~\cite{HoweD02,DBLP:conf/ijcai/VallatiHCM15}. Hence, we will consider \emph{totally ordered sets} 
of elements (with no repetitions) rather than plain sets. Roughly speaking, we formalise, besides the elements themselves, the order in which they are written in a domain description language (e.g., in PDDL).  

\begin{definition}\label{def:conf}
Let $Dom = (P,Ops)$ be a planning domain model. Let $\preceq_P, \preceq_{Ops}$ be total order relations on the sets $P$, $Ops$, respectively. For each $o\in Ops$ let $\preceq_{\MathWord{pre}(o)}, \preceq_{\MathWord{eff}(o)}$ be total order relations on the sets $\MathWord{pre}(o)$ and $\MathWord{eff}(o)$, respectively. For simplicity, we denote a set of the aforementioned relations as $\preceq$.
Then, we say that $Dom^\preceq$ is a \textbf{domain model configuration} of $Dom$ with respect to $\preceq$ 
\end{definition}

Two different domain model configurations of the well known BlocksWorld domain are depicted in Figure~\ref{fig:example}. For instance, the operator total ordering relations $\preceq_{Ops}$ and $\preceq'_{Ops}$ are defined as
\begin{center}
\texttt{pick-up} $\preceq_{Ops}$ \texttt{put-down} $\preceq_{Ops} $\texttt{stack} $\preceq_{Ops} $\texttt{unstack} and\\ \texttt{unstack} $\preceq'_{Ops}$ \texttt{stack} $\preceq'_{Ops} $\texttt{pick-up} $\preceq'_{Ops}$ \texttt{putdown}
\end{center}
for the model configurations in Figure~\ref{fig:example} left and right, respectively. It should be noted that ``elementary'' total order relations are independent, i.e., predicates in a precondition of an operator might not be ordered in the same way as in the predicates definition (as can be seen in both cases in Figure~\ref{fig:example}).

The \emph{domain model configuration space} that consists of all possible domain model configurations (i.e., all possible $\prec$ relations) can be very large even for simple models. Its size is determined by the numbers of permutations of particular sets of elements (e.g., predicates, preconditions) as shown in the following equation:

$$|P|!\,|Ops|!\prod_{o\in Ops}(|\MathWord{pre}(o)|!\,|\MathWord{eff}(o)|!)$$

For illustration, the number of possible configurations of the \texttt{pick-up} operator is only 144 while the size of the BlocksWorld domain model configuration space is already about $1.7\cdot10^{12}$.







\section{Impact of Domain Model Configuration}\label{sec:impact}

This section is devoted to the empirical evaluation of the influence of domain model configuration on the performance of state-of-the-art planning engines, and to the discussion of techniques for compensating this source of performance variation.

Previous work in this area --such as that of Howe and Dahlman \cite{HoweD02}-- has presented evidence that changing the order of elements in either domain models or problem specifications can affect performance. In this section, we aim to analyse how the relative performance of planners changes due to domain model configuration. In fact, this can be seen as a source of performance variation that can affect the outcome of competitions, in terms of final rankings. 

\subsection{Experimental Setup}

For this experimental analysis, in order to assess the impact of randomly configured domain models, we chose to re-run the \emph{Agile} sequential, deterministic track
of the 2014 International Planning Competition \cite{DBLP:journals/aim/VallatiCGMRS15}. In the \emph{Agile} track, competing planners are evaluated based on the running time required to find an initial satisficing solution plan, with no regard to the quality of that plan. Its emphasis on planner running time and low resource requirements made it ideally suited for our analysis.

In the IPC competition, 14 planning domains were considered, namely: \textit{Barman}, \textit{Cave-Diving}, \textit{Child-Snack}, \textit{CityCar}, \textit{Floortile}, \textit{GED}, \textit{Hiking}, \textit{Maintenance}, \textit{Openstacks}, \textit{Parking}, \textit{Tetris}, \textit{Thoughtful}, \textit{Transport}, \textit{Visitall}. For each domain, 20 benchmark problem instances --the most up-to-date version of those used during IPC 2014-- are considered for the evaluation of the planners' performance. In this empirical study we consider only domains in which a single domain model is used, to better isolate the impact of the model's configuration. For this reason, \textit{Openstacks} has been removed; it provides a different domain model for each problem.

In order to empirically assess the influence of differently configured domain models on planning engines, we generated 50 random domain model configurations of each model (according to Definition~\ref{def:conf}). In plain words, we randomly configured the order in which domain predicates are declared and operators are listed in the PDDL domain model and, within every operator, the order in which precondition and effect predicates are shown.

In our experiments, we used 12 planners. There were 15 competing planners in the IPC 2014 \emph{Agile} track, but out of those 15, 3 competitors submitted two versions of the same planner: IBaCoP, Yahsp3, and Madagascar (Mpc). Given the significant resource requirements of running the experiments on 50 different domain models per each benchmark domain, which corresponds approximately to run 50 times the complete IPC Agile track, we decided to consider only one version of the aforementioned planners. Since the software framework exploited by different versions of a planning engine does not significantly change, we believe that the exclusion of one version of a planner does not reduce the generality of our empirical analysis.

Experiments were performed on a quad-core 3.0 Ghz CPU. Each planner was given a running time limit of 300 CPU seconds, and 4GB of RAM.
Performance is measured in terms of IPC score, PAR10 and coverage. We defined IPC score as in the Agile track of IPC 2014: for a planner $\cal C$ and a problem $p$, {\it Score}$({\cal C},p)$ is 0 if $p$ is unsolved, and $1/(1 + \log_{10}(T_{p}({\cal C})/T^*_{p}))$ otherwise (where $T^*_{p}$ is the minimum time required by the compared systems to solve the problem).
The IPC score on a set of problems is given by the sum of the scores achieved on each considered instance. PAR10 (Penalised Average Runtime), is a metric usually exploited in algorithm configuration techniques, where average runtime is calculated by counting runs that did not find a plan as ten times the cutoff time~\cite{HutterHLS09}. 
Intuitively, PAR10 provides a good tradeoff information between runtime and coverage. It should be noted that the IPC score of a given planner is relative to the performance of the other considered planners; instead, PAR10 is absolute and does not take into account the performance of other competitors. 

It is common practice to include randomised components in AI planners. In order to account for this potential source of noise, each experiment has been run three times: results provided correspond to the median performance. 
Where possible, seeds of planners have been fixed. 

\subsection{Experimental Results}

Table \ref{ipc:summary} gives the overall performance, in terms of IPC score and coverage, achieved by considered Agile planners on the IPC 2014 benchmarks. Results shown include the best performance, the worst performance, and the cumulative median of achieved performance.  Best and worst performance have been identified by selecting the domain model configuration allowing a planner to provide its best and worst results on a domain-by-domain basis. The cumulative median value is obtained by summing the median score achieved by a planner on each domain, and gives an idea of how a planner would perform on average on a set of different domains; in other words, the cumulative median is the sum of domain-wise medians. It is worth remarking that planners react very differently to the provided domain configurations. This means that, for instance, the domain model configuration that allows arvandherd to deliver the best IPC score on the Barman domain, is different from the configuration that allows Probe to deliver its best performance in the same domain. 

\begin{table}
\centering
\caption{\label{ipc:summary} Summarisation of the performance gap that can be achieved by running planners on differently configured domain models. Results show the best (B), worst (W), and cumulative median (CM) IPC score and coverage achieved on considered benchmarks from IPC 2014. Cumulative median is calculated by summing domain-wise median values. Planners are ordered according to the cumulative median IPC score. Bold indicate planners with a statistically significant difference between Best and Worst performance.}
\footnotesize
\begin{tabular}{l | ccc|ccc}
&  \multicolumn{3}{c}{IPC score} &   \multicolumn{3}{|c}{\# Solved} \\
& CM & B & W & CM& B & W\\
\hline

Cedalion \scriptsize{\cite{DBLP:conf/aaai/SeippSHH15}}& 82.80 & 84.56 & 81.97 & 144 & 145 & 144\\
\textbf{arvandherd} \scriptsize{\cite{arvandherd}} & 62.94 & 73.48 & 54.87 & 133 & 151 & 121\\
\textbf{Mpc} \scriptsize{\cite{madagascar}} & 62.31 & 73.38 & 58.75 & 89 & 108 & 80\\
\textbf{Jasper} \scriptsize{\cite{jasper}}& 57.53 & 69.97 & 47.65 & 101 & 133 & 96\\
Mercury \scriptsize{\cite{mercury}}& 55.65 & 58.03 & 54.44 & 104 & 107 & 103\\
SIW \scriptsize{\cite{probe}}& 52.49 & 55.86 & 49.60 & 74 & 78 & 72\\
Bfs-f \scriptsize{\cite{probe}}& 51.70 & 58.78 & 46.76 & 93 & 99 & 86\\
\textbf{Probe} \scriptsize{\cite{probe}}& 48.20 & 63.33 & 37.34 & 83 & 104 & 67\\
\textbf{Yahsp3} \scriptsize{\cite{yahsp3}}& 44.64 & 59.14 & 44.05 & 65 & 84 & 64\\
Freelunch \scriptsize{\cite{freelunch}}& 39.84 & 40.26 & 39.00 & 57 & 57 & 56\\
use \scriptsize{\cite{use}}& 37.43 & 46.98 & 35.78 & 72 & 88 & 70\\
\textbf{IbaCoP } \scriptsize{\cite{DBLP:journals/jair/CenamorRF16}}& 28.57 & 35.08 & 28.04 & 74 & 90 & 74\\

\end{tabular}

\end{table}

Results presented in Table \ref{ipc:summary} confirm the significant impact of domain model configurations on most of the state of the art planning engines, leading to some remarkable score fluctuations. In many cases, delivered performance are statistically significantly different, according to the Wilcoxon signed rank test, $p-value = 0.05$ \cite{wilcoxon}. Probe shows the largest score fluctuation, its IPC score ranges between 63 and 37. Significant fluctuations have been observed also in the IPC score of Jasper (69--47), arvandherd (73--54), Madagascar (73--59) and use (47--36). Remarkable fluctuations can also be noticed in coverage results. On the other hand, evidence indicate that many planners based on the Fast Downward \cite{DBLP:journals/jair/Helmert06} framework, such as Cedalion, Mercury, and Freelunch, are usually not strongly affected by the configuration of the domain models. 
This is possibly due to the fact that during the pre-processing step Fast Downward re-orders the operators alphabetically. This arbitrary ordering of the operators 
does not allow to fully exploit the potential speed-up resulting from a smart configuration of the domain model elements.

Besides the top performing planner (Cedalion), it is apparent that competitions ranks are not stable and are significantly affected by the exploited domain model configuration. For example, Probe, which ranked eighth according to cumulative median performance, could obtain the second place with carefully chosen domain models (and poor domain models for some of the other planners), and could drop to the eleventh 
position with adversarially-chosen domain models. Similar rank changes can be observed for other competitors. 

Our intuition behind the substantial performance variations we report with different domain models throughout this paper is that a different ordering can influence tie breaking; for instance, in A* search, situations where several nodes have the same best f-value are common. 
Intuitively, all of these nodes are equally good and many implementations thus select the node that comes first or last. Therefore, a different order of operators/predicates can drastically change the order in which nodes are considered, thus changing the behaviour of the search. Also, a smart ordering of preconditions can lead to performance improvements in cases where some preconditions are more likely to be unsatisfied than others; having them as first in the checking list can avoid a possibly large number of checks. A similar behaviour has been empirically observed by Valenzano, Schaeffer, Sturtevant, and Xie \cite{valenzano2014comparison}, who noticed that changing the order in which operators are considered during search significantly affects the performance of greedy best-first-based planning systems. 
Moreover, we empirically observed that differently-configured domain models can lead to different SAS+ encodings generated by planners built on top of the Fast Downward framework~\cite{DBLP:journals/jair/Helmert06}.

It is important to assess if observed performance variations are due to a small set of domains, or if all the benchmark domains equally affect the planners' behaviour. In order to shed some light on this aspect, domain-by-domain results are provided in Appendix \ref{appendixdomain}. Results are presented in terms of PAR10 and coverage, to isolate the performance of each planner. Unsurprisingly, evidence indicate that the configuration of domain models composed by a single PDDL operator --like Maintenance and Visitall-- has a very limited impact on the planners' performance. 
On the other hand, in most of the domains the configuration of the models led to noticeably different planner's performance. Table \ref{citycar} show how planners are affected by different models of the CityCar domain. 
For some planners, such as use and Jasper, a non-suitable configuration can result in a $0\%$ coverage performance; whereas running on the Best domain model configuration may boost the coverage up to $25\%$. Arguably, for many of the considered planners, the configuration can dramatically modify the way in which the search space is explored. 

\begin{table}
\centering

\caption{\label{citycar} Results show the best (B), worst (W), median (M) PAR10 and coverage performance achieved on the CityCar IPC 2014 benchmarks when running planners on $50$ randomly configured domain models. ``std'' column reports the standard deviations. Planners are ordered alphabetically. }
\footnotesize
\begin{tabular}{l | cccc|cccc}
&   \multicolumn{4}{|c}{PAR10} &  \multicolumn{4}{|c}{\# Solved} \\
&  B & W & M & std  &  B & W & M & std \\
\hline
 \multicolumn{9}{c}{\bf CityCar} \\
\hline
arvandherd  & 203.09 & 788.74 & 498.80   & 161.13  &  19 &  15 &  17 & 1.41  \\
Bfs-f & 2116.23 & 2258.57 & 2253.84 & 41.08 &  6 &  5 & 5 & 0.18\\
Cedalion  & 2137.34 & 2138.13 & 2137.44   & 0.27   &  6 &  6 &  6 & 0.0   \\
Freelunch  & 3000.00 & 3000.00 & 3000.00   & 0.0   &  0 &  0 &  0 & 0.0   \\
IbaCoP  & 1713.18 & 2570.38 & 1997.55   & 223.69  &  9 &  3 &  7 & 2.41   \\
Jasper  & 2258.35 & 3000.00 & 2265.04   & 342.01  &  5 &  0 &  5 & 2.54   \\
Mercury  & 2430.31 & 2702.90 & 2429.63   & 77.40   &  4 &  2 &  4 & 0.15   \\
Mpc  & 1384.71 & 1811.22 & 1532.23   & 112.60  &  11 &  8 &  10 & 1.07   \\
Probe  & 763.78 & 2401.24 & 1662.58   & 373.69   &  15 &  4 &  9 & 2.45   \\
SIW  & 2255.80 & 2255.95 & 2255.87   & 0.04   &  5 &  5 &  5 & 0.00   \\
use  & 2721.95 & 3000.00 & 3000.00   & 73.78   &  2 &  0 &  0 & 0.20   \\
Yahsp3  & 3000.00 & 3000.00 & 3000.00 & 0.0 &  0 &  0 & 0 &  0.00    \\
\hline

\end{tabular}

\end{table}

Counterintuitively, the number of operators does not seem to be directly related to the impact of the domain model configuration. In fact, domain models that include the largest number of operators are not those in which the configuration led to the largest performance discrepancies. This is the case for Thoughtful, which is the domain model with the largest number of operators (more than twenty) among the ones considered. Our conjecture is that the order of operators tends to have the strongest impact on the planners' performance, but its overall impact significantly varies according to the way in which the operators can be used. In domain models like Thoughtful, where each action has a similar probability to be applied regardless of the previously performed action, operators' order has a very limited impact. Instead, in domain models where there is some sort of sequentiality between actions, the way in which operators are listed gains importance. Given the randomised nature of the considered domain models, it is hard to clearly isolate and assess the importance of the order of preconditions and effects. 

Best, worst, and median performance shown in Table \ref{citycar} and Appendix \ref{appendixdomain}, refer to the performance achieved using a single domain model configuration for solving all the benchmarks of the considered domain. Empirical evidence indicates that a planner tends to perform consistently when run on the same domain model configuration, on the IPC benchmarks. In other words, what has been termed as the \textit{best} configuration, usually allows a planner to deliver the best performance on all the benchmarks of the domain. Moreover, we observed that the coverage of a planner using different domain model configurations is not complementary: i.e., it is not possible to improve the coverage of a given planner by considering different domain model configurations on different planning instances. The set of problem instances solved by a planner run on the \textit{best} domain model configuration is a superset of what the planner can solve, using any of the other considered configurations.  
A significant side effect of this behaviour resides in the fact that it is not useful to create portfolios of differently configured domain models. According to our results, it is in fact more efficient to identify a promising configuration, and then use it for solving the considered benchmarks.

\section{Improving Planners' Performance via Automated Configuration of Domain Models}\label{sec:configuration}

In the previous section, we empirically demonstrated that the configuration of domain models differently affect the performance of domain-independent planners: some planners are strongly affected while others do not show significant performance variations. 
The question naturally arises: \textit{Is it possible to configure domain models in order to maximise the performance of state-of-the-art planners?}

This section is devoted to answering this question. Here we consider two approaches for the effective configuration of domain models. The first one, described in Section \ref{ijcai15}, is based on the exploitation of algorithm configuration techniques, and relies on an expensive learning phase. The underlying idea is that it is possible to collect a sufficient number of training instances from a given domain, and exploit them for optimising the configuration of the corresponding domain model, in order to boost the performance of domain-independent planners on previously-unseen instances from the same domain. 
The second approach, presented in Section \ref{socs17}, introduces heuristics for the online configuration of domain models; this sort of approach does not require any training phase, and is therefore designed for scenarios in which training instances are not available, or are extremely expensive to collect.

Next to making a significant contribution to planning speed-up, the investigation on the possibility of configuring domain models in order to improve planners' performance can shed some light into knowledge engineering for planning, and provide useful information to help engineer more efficient planning domain models.

\subsection{Automated Configuration of Domain Models as Algorithm Configuration}\label{ijcai15}


In this section, we explore the {\it automated configuration of PDDL domain models} by using algorithm configuration approaches. We consider as the parameters for algorithm configuration the potential orderings of predicate declarations, the ordering of operators, and the ordering of predicates within preconditions and effects (as in Def.~\ref{def:conf}). 

\subsubsection{Algorithm Configuration}
Most algorithms have several free parameters that can be
adjusted to optimise performance (e.g., solution cost, or runtime to solve
a set of instances). Formally, this \emph{algorithm configuration} problem can be stated as follows: given a
parametrised algorithm with possible configurations $\mathcal{C}$,
a benchmark set $\Pi$, and a performance metric $m(c,\pi)$ that measures the performance of configuration $c\in\mathcal{C}$ on instance
$\pi\in\Pi$, the performance of $c$ is determined as:
\begin{equation}
\label{eq:ac}f(c) = \frac{1}{|\Pi|} \sum_{\pi\in\Pi} m(c,\pi).
\end{equation}

Ideally, we would want to find a configuration $c^\ast\in\mathcal{C}$ such that $f(c^\ast)\leq f(c)$ for every $c\in\mathcal{C}$, but such a configuration may not exist. Using Equation \ref{eq:ac}, we therefore aim to find the configuration with the best expected performance (or, for a finite benchmark set $\Pi$, the average performance). We note, however, that one could also optimize for other objectives.\footnote{We found optimizing for \emph{median} performance to \emph{not} be a good idea, since it can yield configurations that work very well on 51\% of the instances but extremely poorly on others; in contrast, the mean is often quite dominated by the worst cases, and optimizing it therefore also reduces the failure rate. Our performance metric $m$ can also already penalize failures substantially, allowing us to use Equation \ref{eq:ac} to truly minimize the failure rate and only break ties by the average performance in non-failure cases.}

The AI community has recently developed dedicated algorithm configuration systems to tackle this problem~\cite
  {HutterHLS09,AnsoteguiST09,YuanSB10,smac}. Here we employ the sequential model-based algorithm configuration method SMAC~\cite{smac}, which represents the state of the art of configuration tools and, in contrast to ParamILS~\cite{HutterHLS09}, can handle continuous parameters. 
SMAC uses predictive models of algorithm performance~\cite{epms}
to guide its search for good configurations. More precisely, it uses
previously observed $\langle{}$configuration, performance$\rangle{}$ pairs $\langle{}c, f(c)\rangle{}$ and
supervised machine learning (in particular,
random forests~\cite{breiman2001random})
to learn a function $\hat{f}:\mathcal{C} \rightarrow \mathds{R}$ that predicts the
performance of arbitrary parameter configurations (including those not yet evaluated).
The performance data to fit these models is collected sequentially. In a nutshell, after an
initialisation phase, SMAC iterates the following
three steps: (1)~use the performance measurements observed so far to
fit a random forest model $\hat{f}$; (2)~use $\hat{f}$ to select
promising configurations $C_{next} \subset \mathcal{C}$ to evaluate next,
trading off exploration in new parts of the configuration space
and exploitation in parts of the space known to perform well; and
(3)~run the configurations $c \in C_{next}$  on one or more benchmark instances and compare their
performance to the best configuration observed so far.
%
SMAC is an anytime algorithm that interleaves the exploration of new configurations with additional runs of the current best configuration to yield both better and more confident results over time. As all anytime algorithms, SMAC improves performance over time; for finite configuration spaces it is guaranteed to converge to the optimal configuration in the limit of infinite time.

\subsubsection{Configuration of Domain Models Exploiting Algorithm Configuration Approaches}
In this work, we use SMAC for configuring a domain model $\mathcal{M}$ in order to improve the runtime performance of a given planner $\mathcal{P}$. In this setting, it is possible to evaluate the ``quality'' of a domain model configuration $\mathcal{M}^\prec$ of $\mathcal{M}$ with respect to some $\prec$ by the performance of $\mathcal{P}$ when run on problem instances encoded using $\mathcal{M}^\prec$.
Therefore, on a high level our approach is to express the degrees of freedom in the domain model as parameters, and use algorithm configuration to set these parameters to optimise planner performance.

It is not directly obvious how to best encode the degrees of freedom in domain models as parameters.
Configuration over orders is not natively supported by any general configuration procedure, but needs to be encoded using numerical or categorical parameters. 
The simplest encoding for an ordering of $m$ elements would be to use a single categorical parameter with $m!$ values. However, next to being infeasible for large $m$, this encoding treats each ordering as independent and does not allow configuration procedures to discover and exploit patterns such as ``element $i$ should come early in the ordering''. Two better encodings include:
\begin{enumerate}
\item \textbf{$m \choose 2$ binary parameters}: each parameter expresses which of two elements should appear first in the ordering, with the actual order determined, e.g., by ordering elements according to the number of elements that should appear after them. 
\item \textbf{$m$ continuous parameters}: each element $i$ is associated a continuous ``precedence'' value $p_i$ from some interval (e.g., $[0,1])$, and elements are ordered by their precedence. 
\end{enumerate}

To select which of these two strategies to use in our analysis, we ran preliminary experiments to evaluate them; in these experiments, we configured one domain model (Depots) for each of two planners (LPG and Yahsp3). In both of these experiments, the two strategies led to domain model configurations of similar quality, with a slight advantage for the second strategy; we therefore selected that strategy for our study.
Nevertheless, we would like to stress the preliminary nature of our comparison and that one could encode configuration across orders in many other possible ways; we believe that a comprehensive study of the possible strategies may well lead to substantial performance improvements (and would thereby boost domain configuration further). We leave such a study to future work to fully focus on domain configuration here.


We now describe the encoding we chose in more detail. To express domain configuration as an algorithm configuration problem we introduce a numerical parameter $p_i$ (with domain $[0, 1]$) for each configurable element (i.e., each domain predicate, operator, pre-condition and effects of an operator).
Thus, the configuration space for domains with $m$ configurable elements is $\mathcal{C} = [0,1]^m$. 
A configuration $c = \langle p_1, ..., p_m \rangle \in \mathcal{C}$ instantiates each of the parameters, and the corresponding domain model configuration for it is obtained by grouping and sorting the parameters as follows. 
For domains with $k$ operators, PDDL elements are grouped into $2+2k$ groups: predicate declarations, operators and, for each operator, preconditions and effect predicates. Within each group, items are sorted in order of increasing parameter value; ties are broken alphabetically. Following this order, elements are listed in the resulting PDDL model. The simplest domain we consider here is Parking (see Example 1), with 41 parameters. The most complex domain we considered is Rovers, with 109 parameters.

It is easy to notice that the described encoding does come with a significant amount of redundancy. This is because the sum of the values of the $p_i$ are unnormalised (i.e., they do not sum to 1); scaling all of them by a positive factor preserves the same ordering, and therefore there are potentially infinite configurations of parameters that result in the same configuration of the PDDL model. However, we note that if the parameters were normalised, there would also be some redundancy (namely, the last $p_i$ would follow from the others) or forbidden parameter combinations (where, e.g., the sum of the first few parameter values already exceeds 1), or conditionality relationships on the values (the value range for $p_2$ would only be $[0,1-p_1]$) and the order in which $p_1, p_2, ... p_n$ are presented to the optimiser actually matters in this case since the choice of $p_1$ would be important for all value ranges of $p_2, ..., p_n$. This encoding is also not supported by any algorithm configuration procedure we are aware of. As previously mentioned, we are aware that there may be different ways for encoding the configuration of domain models. Again, we leave this study to future work.

\begin{exmp}
The simplest domain model we consider here, Parking, contains 5
domain predicates and 4 operators, with 3, 4, 3, and 4 pre-conditions and 4, 5, 5, and 4 effects, respectively. We need to specify 10 different orders: the order of the 5 domain predicates; the order of the 4 operators; 4 pre-condition orders (one per operator); and 4 effect orders (one per operator).   
Our formulation of the domain configuration problem for the Parking model has 41 parameters with domain $[0,1]$: 5 domain predicate precedence parameters, 4 operator precedence parameters, (3+4+3+4) pre-condition precedence parameters, and (4+5+5+4) effect precedence parameters. For the sake of simplicity, in the remainder of this example, we focus on configuring only the degrees of freedom of the first operator. It has 3 pre-conditions $\{pre1$, $pre2$, $pre3\}$ and 4 effects $\{eff1$, $eff2$, $eff3$, $eff4\}$. This gives rise to 7 parameters: $p1$, $p2$, $p3$ (which control the order of pre-conditions) and $p4$, $p5$, $p6$, $p7$ (which control the order of effects). SMAC therefore searches in the parameter space $[0,1]^7$, and each parameter setting induces a domain model. If SMAC evaluates, e.g., [$p1$=0.32, $p2$=0.001, $p3$=0.7, $p4$=0.98, $p5$=0.11, $p6$=0.34, $p7$=0.35], this induces a domain model in which preconditions are ordered $pre2\prec_{\MathWord{pre}} pre1 \prec_{\MathWord{pre}} pre3$), and effects are ordered $\MathWord{eff2}\prec_{\MathWord{eff}} \MathWord{eff3}\prec_{\MathWord{eff}} \MathWord{eff4}\prec_{\MathWord{eff}} \MathWord{eff1}$. 
%
\end{exmp}

%
%

A major difference between standard algorithm configuration on the one hand and domain configuration on the other is that the former only applies to planners that expose a large number of explicit tuning parameters. For this reason, within planning, standard algorithm configuration has only been applied to LPG~\cite{VallatiFGHS13} and Fast Downward~\cite{fdauto}. In contrast, in domain configuration we configure over the \emph{inputs} of a planner, which makes this technique planner-independent.


\subsubsection{Settings}


For this analysis, we selected 6 planners, based on their performance in the Agile track of the international planning competition (IPC) 2014 
and/or the use of different planning approaches: Jasper \cite{jasper}, Lpg \cite{lpg}, Madagascar (Mpc) \cite{madagascar}, Mercury \cite{mercury}, Probe \cite{probe}, and Yahsp3 \cite{yahsp3}. The aim is to consider a set of very different approaches to planning, in order to gain a better understanding of the possible improvement that can be gained by specifically configuring domain models. 



No   portfolio-based planners were included, since the exploitation of different planning techniques on the same planning instance can possibly mitigate the usefulness of domain model configuration. This was driven by our intuition that domain configuration is a planner-specific tuning.

We focused our study on domains that have been used in IPCs and for which a randomised problem generator is available that allows to generate the number of problems needed for training. 
The models we chose had at least four operators and, possibly, a large number of predicates:
Blocksworld (4 ops version), Depots, Matching-Bw, Parking, Rovers, Tetris, and ZenoTravel. 

For each domain we study, we created approximately 550 random instances with the domain's random instance generator.
We split these instances into a training set (roughly 500 instances) and a test set  (roughly 50 instances) in order to obtain an unbiased estimate of generalisation performance to previously unseen instances from the same distribution. Throughout the paper, we only report results on these test sets.

Configuration of domain models was done using SMAC version 2.08 (publicly available at \url{http://www.aclib.net/SMAC}).
As in Section \ref{sec:impact}, the performance metric we consider and minimise is penalised average runtime (PAR10).
%

We performed experiments on AMD Opteron\texttrademark{} machines with 2.4 GHz, 8 GB of RAM and Linux operating system. Each of our configuration runs was limited to a single core, and was given an overall runtime and memory limits of 5 days and 8GB, respectively. As in the Agile track of the IPC 2014, the cutoff time for each instance, both for training and testing purposes, was 300 seconds. 

\begin{table}[tp]

\scriptsize

  \begin{center}
          \caption{\label{tab:exp1} Test set results in terms of PAR10, percentage of solved problems and IPC score, of planners running on original domain models (O) and SMAC-configured models (C). Between brackets the number of testing instances considered for the domain. Boldface indicates statistically significant improvements of PAR10 performance.
        IPC score was calculated separately for each planner, considering only the performance of its two domain model configurations. 
        The decrease in IPC score of Yahsp3 on Matching-Bw by configuration is not a typo: even though the optimisation objective (PAR10) substantially improved, it did so at the cost of solving easy instances more slowly. 
}
    \begin{tabular}{l|cc|cc|cc}
    & \multicolumn{2}{|c|}{PAR10} & \multicolumn{2}{|c|}{Solved} & \multicolumn{2}{c}{IPC score} \\
   
 Planner & C & O & C & O & C & O\\
\hline
\multicolumn{7}{c}{\bf BlocksWorld (55) }  \\
\hline
Jasper &{\bf 307.3}&470.1&90.9& 85.4  & 49.0 &45.0\\ 
Lpg &{\bf 5.9}&10.3&100.0& 100.0  & 53.4 &45.1\\ 
Mpc &2896.0&2949.6&3.6& 1.8 &2.0&0.9\\
Mercury &90.6&90.3&98.2& 98.2  &53.5&53.8\\
Probe &{\bf 29.3}&30.4&100.0& 100.0  & 54.7 &53.9\\ 
Yahsp3 &2.9&57.5&100.0&  98.2 & 47.4&47.8\\
\hline
\multicolumn{7}{c}{\bf Depots (49) }  \\
\hline
Jasper &2295.0&2241.3&26.4& 24.5&  12.9&11.6\\
LPG &{\bf 9.0}&78.3&100.0& 98.0& 47.0&40.9\\
Mpc &3.7&3.7&100.0& 100.0& 48.0&47.9\\
Mercury &2883.4&2883.4&4.1& 4.1& 2.0&2.0\\
Probe &14.1&14.2&100.0& 100.0& 47.8&47.4\\
Yahsp3 &{\bf 1.1}&373.4&100.0& 87.8&  49.0 &31.7\\
\hline
\multicolumn{7}{c}{\bf Matching-Bw (50)}  \\
\hline
Jasper &524.1&532.1&84.0& 84.0 &38.7&37.9\\
LPG &{\bf 685.2}&1278.0&78.0& 58.0& 35.8 &22.9\\
Mpc &3000.0&3000.0&0.0& 0.0& -- & -- \\
Mercury &1695.3&1694.5&44.0& 44.0&21.7&21.8\\
Probe &1464.0&1338.6&52.0& 56.0&23.3&25.8\\
Yahsp3 &{\bf 18.5}&185.7&100.0& 94.0& 34.9 &44.2\\
\hline
\multicolumn{7}{c}{\bf Parking (56)} \\
\hline
Jasper &1446.2&1448.9&53.6& 53.6&29.8&29.2\\
LPG &2947.0&2846.1&1.8& 5.4& 1.0&2.7\\
Mpc &855.3&858.1&73.2& 73.2 &39.8&38.9\\
Mercury &1446.2&1444.8&53.6&53.6&29.7&29.9\\
Probe &1268.0&1418.1&58.9& 53.6 &29.6&28.1\\
Yahsp3 &2059.8&1955.1&32.1& 35.7&16.8&19.6\\
\hline
\multicolumn{7}{c}{\bf Rovers (60)}  \\
\hline
Jasper  &{\bf 2498.6}&2544.1&18.3& 16.7&11.0&9.9\\
Lpg  &{\bf 23.7}&26.1&100.0& 100.0& 59.1 &56.9\\
Mpc &3000.0&3000.0&0.0& 0.0& -- & -- \\
Mercury  &{\bf 1173.4}&1536.9&66.7& 53.3& 40.0 &31.7\\
Probe  &3000.0&3000.0&0.0& 0.0 &-- &-- \\
Yahsp3  &7.5&7.5&100& 100.0&59.9&59.8\\
\hline
\multicolumn{7}{c}{\bf Tetris (54)} \\
\hline
Jasper &1937.3&1944.1&37.0& 37.0&19.5&18.9\\
LPG &{\bf 1907.8}&2723.1&37.0& 9.3&  19.8&3.6\\
Mpc &1860.9&1863.3&38.9& 38.9&21.0&20.3\\
Mercury &1985.1&1983.4&35.2& 35.2&18.6&18.9\\
Probe &{\bf 1490.6}&2520.5&51.9& 16.7& 28.0 &4.5\\
Yahsp3 &2615.8&2614.6&13.0& 13.0&6.4&7.0\\
\hline
\multicolumn{7}{c}{\bf ZenoTravel (50)}  \\
\hline
Jasper &{\bf 174.9}&251.4&100.0& 98.0 & 50.0 &46.8\\
LPG &{\bf 36.4}&43.1&100.0& 100.0& 49.1 &46.0\\
Mpc &9.5&9.4&100.0& 100.0&49.2&49.1\\
Mercury &84.6&{\bf 82.4}&100.0& 100.0&49.4 &49.9\\
Probe &138.9&143.1&100.0& 100.0&49.1&48.3\\
Yahsp3 &2.5&2.8&100.0& 100.0&47.9&46.1\\

\hline

    \end{tabular}
  \end{center}

\end{table}

\subsubsection{Results}\label{sec:res}

Table \ref{tab:exp1} compares the results of planners running on the original domain model -- the one that has been used in IPCs -- and the specifically configured domain model, obtained by running SMAC. 
These results indicate that the configuration of the domain model has a substantial impact on the planners' performance, with several substantial and statistically significant improvements (as judged by a Wilcoxon signed rank test \cite{wilcoxon}).

\begin{figure}[t]
\centering
  \centering
  \includegraphics[width=0.4\linewidth]{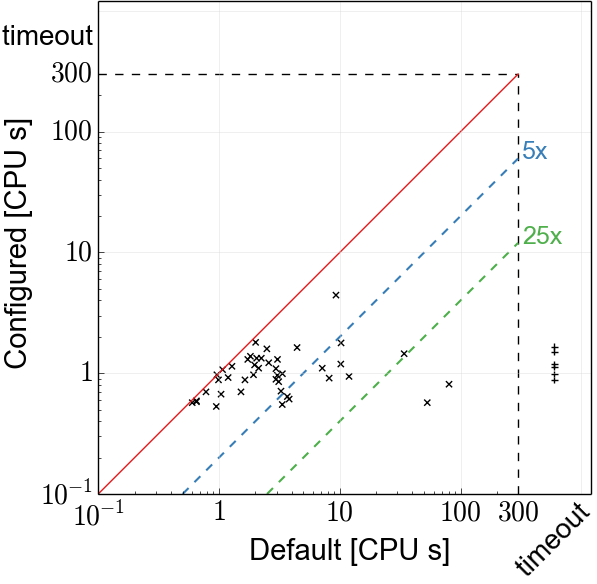}
  \includegraphics[width=0.4\linewidth]{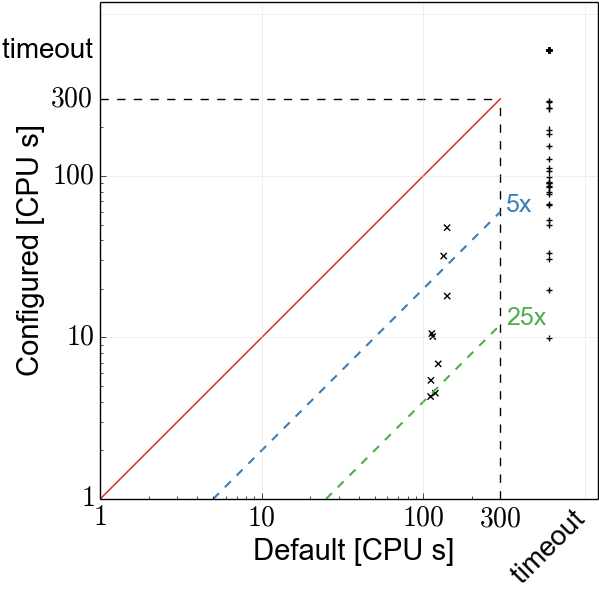}
\caption{\label{figure:sc} Scatter plots comparing runtime of planners using configured (y-axis) and original (x-axis) domain models; timeouts at 300s are plotted separately, dashed lines indicate 5-fold and 25-fold speedups, respectively. The left plot shows the performance of Yahsp3 on benchmarks from the Depots domain. The right plot presents the results of Probe in the Tetris domain.}
\end{figure}

Figure \ref{figure:sc} visualises the improvements for two cases. The left part of Figure \ref{figure:sc} shows that Yahsp3 had mediocre performance on Depots using the original domain configuration (e.g., timing out on 12\% of the instances after 300 seconds), but that after domain configuration it could solve every single instance in less than 10 seconds. The second graph presented in Figure \ref{figure:sc} shows that Probe benefited substantially from domain configuration on the Tetris domain, with both its coverage tripling and its runtime decreasing on solved instances.


Finally, of note is that in three domains domain configuration changed what is the best planner (according to PAR10):
\begin{itemize}
\item In \emph{Blocksworld}, LPG performed best on the original domain model, but its penalised average runtime was ``only'' halved by automatic domain configuration, whereas Yahsp3 was sped up by a factor of 30, ending up beating LPG.
\item In \emph{Depots}, Mpc performed best on the original domain model, but 
since automatic domain configuration helped Yahsp3 reduce its penalised average runtime by a factor of more than 300 (and did not improve Mpc), Yahsp3 clearly performed best on this domain after configuration.
\item In \emph{Tetris}, a similar situation occurred: Mpc performed best on the original domain model, but since automatic domain configuration helped Probe solve three times more problems (and did not improve Mpc), Probe clearly performed best on this domain after configuration.
\end{itemize}

\begin{table}[t]


  \begin{center}
          \caption{ \label{tab:confM} Performance of each planner (one row per planner) running on the domain configurations configured by SMAC (each column represents the domain model configured for one planner) and on the original domain model.  
Respectively, columns Js, LPG, Mp, Me, Pb and Y3 indicate the domain-model configurations identified by SMAC for the Jasper, LPG, Madagascar, Mercury, Probe and Yahsp3 planner. Or indicates the original model. Performance is given as cumulative across all the test instances. IPC score is evaluated by considering all the planners and all the domain model configurations at the same time. Bold indicates best performance, also considering hidden decimals, with respect to each planner.}
    \begin{tabular}{l|c|cccccc}
    
Planners    & \multicolumn{7}{|c}{PAR10} \\ 
   
 & Or & Js & LPG & Mp & Me & Pb & Y3 \\
\hline
Jasper & 1372.3 & {\bf 1293.2} & 1376.9 & 1318.2 & 1338.7 & 1360.7 & 1354.9\\
LPG & 923.7 & 867.9 & {\bf 701.4} & 893.0 & 918.9 & 894.6 & 757.2 \\
Mp & 1799.3 & 1833.3 & 1818.3 & 1802.6 & 1796.6 & {\bf 1781.1} & 1796.2 \\
Mercury & 1377.9 & 1327.1 & 1341.8 & 1349.0 & {\bf 1312.7} & 1320.0 & 1319.9 \\
Probe & 1062.0 & 672.4 & 593.3 & 707.6 & 588.3 & {\bf 578.9} & 619.8\\
Yahsp3 & 389.6 & 311.4 & 319.6 & 391.0 & 352.5 & 326.2 & {\bf 309.8} \\

\hline
\hline

Planners    &\multicolumn{7}{|c}{\% Solved Problems} \\

 & Or & Js & LPG & Mp & Me & Pb & Y3 \\
\hline
Jasper & 56.4 & {\bf 59.1} & 56.1 & 58.3 & 57.5 & 56.7 & 57.0 \\
LPG & 69.8 & 71.7 & {\bf 77.3} & 70.9 & 70.1 & 70.9 & 75.4 \\
Mp & 41.0 & 39.3 & 39.8 & 40.4 & 40.6 & {\bf 41.2} & 40.6 \\
Mercury & 57.0 & 57.8 & 57.2 & 57.0 & {\bf 58.3} & 58.0 & 58.0 \\
Probe & 70.8 & 78.9 & 81.6 & 77.8 & 81.8 & {\bf 82.1} & 80.7 \\
Yahsp3  & 87.4 & 90.1 & 89.8 & 87.4 & 88.8 & 89.6 & {\bf 90.1}\\

\hline
\hline

Planners    &\multicolumn{7}{|c}{IPC score} \\

 & Or & Js & LPG & Mp & Me & Pb & Y3 \\
\hline
Jasper & 86.9 & {\bf 90.6} & 89.0 & 87.8 & 88.6 & 86.8 & 86.5\\
LPG & 163.5 & 170.3 & {\bf 192.6} & 167.2 & 164.8 & 167.5 & 183.9\\
Mp & 103.2 & 100.1 & 100.4 & {\bf 104.4} & 102.7 & 103.3 & 102.2\\
Mercury & 87.9 & 90.3 & 89.5 & 89.1 & {\bf 91.1} & 90.7 & 90.6\\
Probe  & 125.9 & 147.4 & 156.8 & 142.5 & 154.2 & {\bf 158.1} & 152.2\\
Yahsp3  & 277.0 & 285.6 & 285.9 & 272.0 & 259.6 & 285.0 & {\bf 292.3}\\

 \hline

    \end{tabular}
  \end{center}
 
\end{table}

For providing a better overview of the impact of different domain model configurations on the considered planners, we ran all planners on all domain model configurations identified by SMAC. Table \ref{tab:confM} shows the results of this comparison. Generally speaking, performance of planners varied across these domain models, but not substantially. This indicates that there possibly exists a single configuration that can boost performance of all the considered planners. As could be expected, almost all the planners achieved best performance when using the domain model specifically configured for them. Running on domain models configured for different planners tend to lead to worse performance.
The only exception is Mpc, which usually performs slightly better, in terms of PAR10 and solved problems, when running on domain models configured for Probe. 
We hypothesise this is due to the fact that Mpc timed out on more than half the training instances. This slows down the configuration process, likely causing SMAC (in its fixed budget of five CPU days) to only find configurations that are quite far from optimal for this solver.


Next, in order to study how much average impact a change of domain configuration can have across planners, for each domain we identified the ``most improving configuration'', as the configuration that most improved the IPC score of a specific planner. 
(For instance, in Blocksworld the most improving configuration is the one for LPG, which allows the planner to increase its IPC score by $8.3$ points, see Table \ref{tab:exp1}). In Table \ref{tab:gen2} we report the performance of the six planners exploiting, for each domain, the most improving configuration. 
Interestingly, exploiting these configurations allows all the considered planners to achieve statistically significantly better performance than when using the original configuration. This indicates that the original domain models, i.e. the models which are currently used when benchmarking planning systems, are not in a ``planner-friendly" shape. This supports the intuition that there are knowledge engineering policies for encoding domain models that can lead to an average performance improvement of planning systems. 

\begin{table}[t]


  \begin{center}
   \caption{\label{tab:gen2} Results on all the testing instances, in terms of IPC score, PAR10 and percentage of solved problems, of planners running on original domain models (O) and the most improving configured models (B). Bold indicates statistically different IPC score (and PAR10) performance.}
    \begin{tabular}{l|cc|cc|cc}
    & \multicolumn{2}{|c|}{IPC score} & \multicolumn{2}{|c|}{PAR10} & \multicolumn{2}{c}{Solved} \\
   
 Planner &  O & B & O & B & O & B\\
\hline
Jasper &195.9& {\bf 252.2} &1375.9&975.3& 56.3&69.4\\ 
LPG &192.1& {\bf 261.2} &926.2&826.0& 69.7&72.9\\ 
Mpc &167.9& {\bf 232.5} &1637.5&1109.4& 46.1&63.8\\ 
Mercury &183.9& {\bf 236.8} &1381.5&1126.6& 55.8&63.5\\ 
Probe &205.5& {\bf 294.2} &1265.3&602.3& 58.7&81.0\\ 
Yahsp3 &260.0& {\bf 310.9} &390.6&310.7& 87.4 &90.1\\ 
\hline
    \end{tabular}
  \end{center}

\end{table}


Finally, we selected the two domains in which the planners' performance mostly improved by using configured domain models: Tetris and Depots (see Table \ref{tab:exp1}). On these domains, we also used SMAC for identifying a ``bad'' domain configuration, i.e., the configuration that \emph{maximises} PAR10. Maximising PAR10 means identifying configurations that slow down the planning process, in order to increase the number of timeouts. We assessed how such configurations affect planners by comparing performance planners achieved when exploiting them, with those they obtained on the Original and Best models. We observed a noticeable --and statistically significant-- deterioration in performance both in terms IPC score ($-21.1$ vs Original, $-64.4$ vs Best), solved problems ($-2.0\%$ vs Original, $-6.6\%$ vs Best) and average runtime on solved instances ($+4.4$ vs Original, $+16.3$ vs Best) among all the planners. Results are averages across both domains.





\subsubsection{Impact of Configured Models on Relative Planners' Performance}

In order to understand how the different automatically configured models affect the relative performance of planners, we simulate the execution of three ``planning competitions" between the six considered planners on the 50 testing instances of the selected domains. We used the same settings as for the Agile track of IPC-14. In the first competition, planners ran on the original domain models. In the second, all the planners ran on the most improving domain model configurations (Best), selected as described in Section \ref{sec:res}. Finally, in the third competition, every planner ran on the specifically configured domain models (Configured). Although the third competition is unusual, in the sense that planners are actually running on different models, it can provide some interesting insights. Like in the Configurable SAT Solver Competition (CSSC;~\cite{cssc}), it measures the performance the various planners have \emph{when combined with an automated configuration process} (in this case, for configuring the domain).

%
The results of the three competitions are shown in Table \ref{tab:exp2}. Interestingly, the IPC score of all the planners increases when considering Configured models, and it usually improves further on the Best possible model. This is possibly due to the fact that good configurations are similar among considered planners. Moreover, for planners which did not solve many of the training problems, SMAC probably was not able to explore large areas of the configuration space; therefore the specifically configured domain models can be of lower quality than the most improving model. Finally, we notice that in Table \ref{tab:exp2}, the relative performance of planners, i.e. their ordering, 
changed in the three different settings of the competition.
We point out that: (i) IPC score of all the planners increases when considering Configured models, and it usually improves further on the Best possible model; (ii) the relative performance of planners, i.e. their ordering, significantly changes. However, the latter should come as no surprise for the reader, given the results of the analysis performed in Section \ref{sec:impact}.


In order to shed some light on the importance of elements' order, thus providing some useful information for knowledge engineers, we assessed the importance of parameters in two domains: Depots and Tetris. They have a very different structure, and domain configuration substantially improved Yahsp3 and Probe, respectively (compare Figure \ref{figure:sc}).

\begin{table}[t]


  \begin{center}
              \caption{\label{tab:exp2} Relative position of considered planners (IPC score) on original domain models (Original), configured for each planner (Configured) and the configuration that provides the best improvements among all the considered (Best). Planners are listed in decreasing order of IPC score.}
    \begin{tabular}{l|c | c | c}
\hline
Pos & {\bf Original} & {\bf Configured} & {\bf Best} \\
\hline
1&Yahsp3 (297)& Yahsp3 (303)& Yahsp3 (306)\\
\hline
2&Lpg (168)& Lpg (198)& Lpg (189)\\
\hline
3&Mpc (121)& Probe (123)& Probe (173)\\
\hline
4&Probe (112)& Mpc (117)& Jasper (123)\\
\hline
5&Mercury (91)& Mercury (95)& Mpc (119)\\
\hline
6&Jasper (90)& Jasper (94)& Mercury (105)\\
\hline
    \end{tabular}

  \end{center}

\end{table}

\begin{figure}[t]
\centering
\includegraphics[width=.6\linewidth]{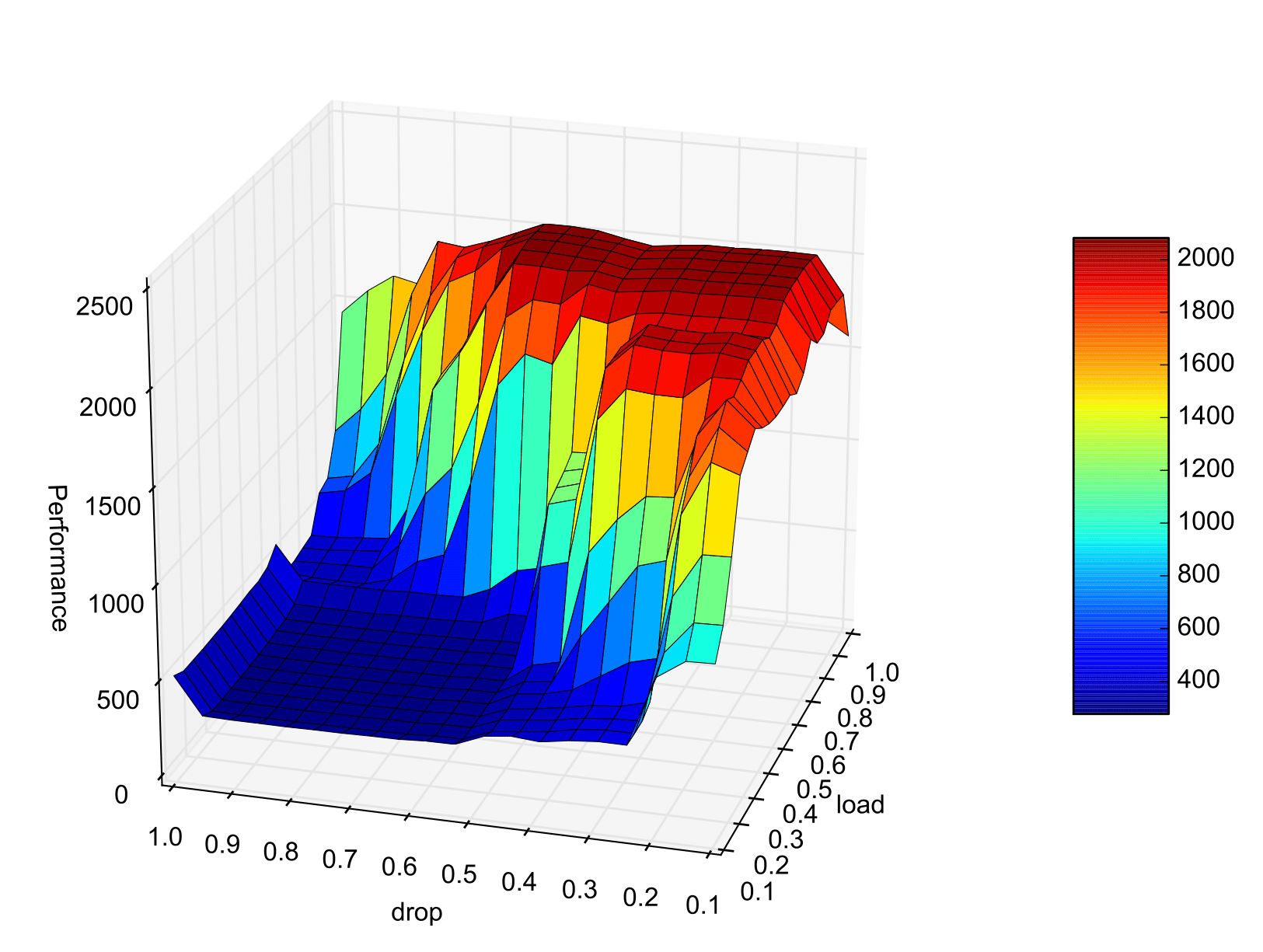}
\caption{\label{fig:opX} The average PAR10 performance of Yahsp3 in the Depots domain, as a function of the values of the parameters corresponding to the {\tt load} 
and {\tt drop} 
operators.
The lower the PAR10 value, the better the performance. Best performance is achieved when the {\tt load} operator is listed early in the model (low parameter value), and the {\tt drop} operator is listed late (high parameter value).} 
\end{figure}

We used fANOVA, a general tool for assessing parameter importance~\cite{hutter2014efficient} after algorithm configuration,
to find that different parameters are indeed important for these two domains.
Although we believed that the order of operators may have the strongest impact on performance, the experimental analysis demonstrated that this may not always be true. 
%
In the two domain models, the order of operators had different impact. In Depots, which has a strong {\it directionality} (some actions are very likely to be executed in initial steps and not in final steps, and vice-versa), ordering operators according to this directionality clearly improves planning performance. In particular, our analysis highlighted that two operators have a very significant impact on performance: {\tt load} and {\tt drop}. Figure \ref{fig:opX} shows a clear pattern for Yahsp3's performance as a function of two parameters in the domain model: 
in order to achieve a low penalised average runtime, operator {\tt load} (which is mostly used in early steps for loading crates on trucks) should be listed early in the model description and operator {\tt drop} should be listed late.

In the Tetris domain, where all the operators have potentially the same probability to be executed in any part of the plan, a good operators' order follows the frequency of action usage in the plan: most used first. Clearly, the frequency of action usage depends, to some extent, on the planner. Nevertheless, in domains such as Tetris, it is likely that the operators' ordering has a less significant impact than in domains with a strong directionality. In line with this explanation, the performance improvements for Tetris arose not by a single factor but as gains due to many of the configurable elements. 

According to our observations in both Depots and Tetris domains, preconditions of operators which are most likely to be unsatisfied should be listed earlier. On the other hand, the impact of ordering of effect predicates is not very clear; our intuition is that the main effect for which an action is usually executed should be presented earlier. Finally, the configured order of predicate declarations tends to follow their frequency as operators' preconditions.

%

Since the results in Table \ref{tab:confM} show that LPG tends to benefit most from domain configuration, we studied its performance in some more detail. In particular, we focused on the Tetris domain, where its percentage of solved instance was increased most.
For gaining some insights, we qualitatively compared the Tetris domain model configured for LPG, with the models configured for other planners and with the aforementioned ``bad'' configuration. 
In terms of predicates, we notice that the {\tt connected} predicate has a strong impact: it is the static predicate that describes connections between cells. Although static predicates are commonly used for describing aspects of the world that do not change, and are included also in some of the other domains considered in our empirical analysis, the {\tt connected} predicate used in Tetris is peculiar. Every operator incorporates at least two preconditions that are based on this predicate. 
LPG performance is improved when {\tt connected} is the last predicate to be defined, but it is listed early as preconditions. 
In terms of operators, those which are rarely used by LPG should be the listed last. 





\subsection{Heuristic Configuration of Domain Models}\label{socs17}

According to the results presented in Section \ref{ijcai15}, the order in which operators are listed in domain models can have a significant impact on the performance of domain-independent planners. Specifically, it has been observed that operators that are often used in solution plans should be listed first in the domain model, while less frequently used operators are better placed last in the model. 

Here we propose heuristics for ordering operators in PDDL models. As we consider the typical domain-independent scenario, where configuration should be performed online, the focus on operators provides a good trade-off between the additional overhead and the potential impact on performance. The underlying hypothesis is that the way in which operators are encoded carries some knowledge about their usefulness and exploitation during the planning process, that should be reflected in their ordering.  

Formally, given a planning domain model $\mathcal{M}$, and the corresponding set of operators $Ops = (o_1, ..., o_m)$, we propose heuristics that, by considering some aspects of the operators in $Ops$, provide as output a total order relation $\prec_{Ops}$. Operators are then listed in the domain model according to $\prec_{Ops}$. No other elements of domain models are modified by the heuristics. 

We introduce five heuristics for ordering operators, that consider the following aspects:

\begin{itemize}

\item \textbf{EFF}. The number of effects.

\item \textbf{PRE}. The number of preconditions.

\item \textbf{RAT}. The ratio between effects and preconditions.

\item \textbf{NEG}. The number of negative effects.

\item \textbf{PAR}. The number of parameters.

\end{itemize}


These considered aspects are quick to compute and can provide an intuition about the expected use of operators. For instance, the presence of a large number of negative effects implies that the corresponding actions are strongly affecting the environment and could therefore be an indication that they are rarely used. Intuitively, such actions may allow to achieve some complex goals or facts, but preventing the subsequent achievement of other goals or facts (or even leading to dead ends). 
On the contrary, the presence of very few preconditions can be an indication of actions that are often used, as the required precondition can be easily satisfied. The ratio between effects and preconditions can give some further insights by considering both aspects at the same time: a high ratio points to actions that have many effects and few preconditions and can thus be used often; a low ratio may denote 
actions that require many preconditions to be satisfied and have a limited impact on the world --although this limited impact may be of critical importance for achieving goals. 
Finally, the number of parameters is an indicator of the expected number of grounded actions. 

Each heuristic has two possible instantiations: ordering operators according to decreasing or increasing values of the considered metric. Hereinafter, we will use numbers to refer to the ordering, and letters for identifying the heuristic. For instance \textbf{EFF1}  indicates that operators are ordered decreasingly with regards to the number of effects, i.e., the operator with the largest number of effects is listed first, and the operator with the least number of effects is listed last. On the contrary, \textbf{EFF2} refers to operators that are ordered \emph{increasingly} according to the number of effects: the first listed operator is the one with the least number of effects. Given the five considered heuristics, and the two possible ordering of each of them, we are able to generate ten different domain models, where operators are ordered in different ways. 

In our implementation, ties are broken following the relative order of operators in the original PDDL model. In terms of complexity, ordering operators according to the proposed heuristics correspond to sorting a list of elements with assigned values. Given the number of operators in a typical PDDL models, ordering time is negligible.

\subsubsection{Experimental Setup}

We selected all the planners that took part in the Agile track of IPC 2014~\cite{DBLP:journals/aim/VallatiCGMRS15} that did not exploit a portfolio-based approach: Arvandherd \cite{arvandherd}, Bfs-f \cite{probe}, Freelunch \cite{freelunch}, Jasper \cite{jasper}, Madagascar (Mpc) \cite{madagascar}, Mercury \cite{mercury}, Probe \cite{probe}, SIW \cite{probe}, use \cite{use}, and Yahsp3 \cite{yahsp3}. As in the previous set of experiments, we did not include portfolio-based planners due to the intuition that the exploitation of different planning techniques on the same planning instance can possibly add noise to the measurement of the usefulness of domain model configuration.

We focused our study on the domain models used in the Agile track of IPC 2014: Barman, Cave-Diving, Child-Snack, CityCar, Floortile, GED, Hiking, Parking, Tetris, Thoughtful, and Transport. Maintenance, Visitall and Openstack domains were not considered for the following reasons. Maintenance and Visitall have a model composed by only one operator, and our proposed heuristics thus would not change the original domain model.
Openstack has a peculiar structure, in the sense that it has a different model per each problem, where elements of problem and domain models are mixed; this can add noise to the empirical evaluation of the effectiveness of proposed heuristics on domain models that are shared among different problems.  

Performance is measured in terms of IPC runtime score, PAR10 and coverage. 
%
Experiments were performed on a quad-core 3.0 Ghz CPU, with 4GB of available RAM. In order to account for randomised algorithms and noise, results provided are averaged across three runs. Where possible, seeds of planners have been fixed. As in the Agile track of IPC 2014, the cutoff time for each instance was 300 seconds. 

\subsubsection{Experimental Results}

Table \ref{tab:h1} shows the cumulative performance, among all the considered domains, of planners running on the original domain model (O) and on the domain model, configured using the considered heuristics, that allows to achieve the best PAR10 score (B). 
For each planner, the best domain model configuration has been selected on a domain-by-domain basis. We noticed that no significant further improvement is gained when configurations are selected on an instance-by-instance basis. 
In other words, performance of planners running on a given domain model configuration is consistent among benchmark instances from the same distribution.

\begin{table}[t]


  \begin{center}
          \caption{\label{tab:h1} Performance, in terms of PAR10, IPC score and coverage, achieved by the considered planners running on original domain models (O) and on the best model configured using the proposed heuristics. The best configured model is selected on a domain-by-domain basis. Bold indicates improvements of more than 10\% on the corresponding metrics. IPC score was calculated separately for each planner; therefore, it cannot be compared across planners.
}
    \begin{tabular}{l|cc|cc|cc}
    & \multicolumn{2}{|c|}{PAR10} & \multicolumn{2}{|c|}{IPC Score} & \multicolumn{2}{c}{Solved (\%)} \\
   
 Planner & O& B & O& B & O & B\\
\hline

Bfs-f & 1977.1 & 1871.3 & 71.6 & \textbf{80.5} & 34.5 & 38.2 \\ 
Mpc & 1994.8 & \textbf{1740.4} & 64.1 & \textbf{83.0} & 34.1 & \textbf{43.2} \\ 
Yahsp3 & 2256.6 & \textbf{2012.4} & 46.0 & \textbf{67.7} & 25.0 & \textbf{33.2} \\ 
Probe & 1876.3 & \textbf{1561.3} & 73.6 & \textbf{93.4} & 38.2 & \textbf{48.7}\\ 
arvandherd & 1350.4 & 1278.8 & 104.6 & \textbf{116.8} & 57.3 & 59.1 \\ 
Freelunch & 2416.0 & 2415.6 & 43.0 & 44.0 & 20.0 & 20.0 \\ 
Mercury & 1771.0 & 1770.6 & 90.9 & 91.4 & 41.4 & 41.8 \\ 
Jasper & 1674.4 & 1582.8 &  86.6 & \textbf{94.8} & 45.0 & 47.3 \\ 
SIW & 1935.5 & 1894.9 & 77.0 & 80.0 & 36.4 & 36.8 \\
use & 1469.8 & \textbf{1300.5} & 100.5 & \textbf{112.5} & 51.7 & 57.3 \\

\hline
    \end{tabular}
  \end{center}

\end{table}

Remarkably, for many of the considered planners, the IPC score can be increased by more than 10\% by choosing the best orders of operators out of those generated by the considered heuristics. 
All planners we considered except Freelunch show an increased percentage of solved instances when running on the best heuristically-configured models. Mpc, Yahsp3, and Mercury 
show statistically significant PAR10 performance improvements, according to the Wilcoxon signed rank test. Noteworthy, Mercury shows a very limited PAR10 and IPC score improvement, but the use of the best configured domain models allows the planner to be consistently --but not dramatically-- faster on most of the benchmarks, when compared to the performance achieved on the original domain model. 
On the contrary, the performance of Freelunch is unaffected by the configuration of the domain model. Even the exploitation of specifically configured models, obtained through the approach introduced in Section \ref{ijcai15}, does not lead to any noticeable performance improvement. Possible reasons for this include the special 
SAT encoding used by Freelunch, the relaxed exist step encoding \cite{balyo2013relaxing}, in which a large number of actions are packed together in a single planning step, as well as the fact that Freelunch has been developed using the Fast Downward framework \cite{DBLP:journals/jair/Helmert06}. Apparently, the order in which operators are listed in the domain model does not affect the shape of the generated SAT formula enough to have an impact on planning performance. 

In order to check how the results presented depend on different cutoff times, we considered longer and shorter thresholds. When a longer runtime of 600 CPU seconds is allowed instead of 300 CPU seconds, the qualitative results do not change much: even though more benchmarks are solved by most of the planners, the relative improvement (percentage) is very similar. However, when only a shorter runtime of 60 CPU seconds are allowed, we observed very different behaviour: the performance of the planners on all the considered models became very similar. 
Our intuition is that on very ``simple'' problems, the way in which the domain model is presented to the planner has a limited importance, as the search space to visit is restricted. On more complex (and larger) instances, instead, the importance of domain model configuration arises, and the gap between ``good'' and ``bad'' heuristics widens.

Results presented in Table \ref{tab:h1} indicate that the designed heuristics are able to provide domain model configurations that positively affect domain-independent planners' performance. This confirms our initial hypothesis, i.e. that the way in which operators are encoded can provide valuable knowledge about the expected use of corresponding grounded actions in solution plans.

\begin{figure}[t]
  \centering
    \includegraphics[scale=0.7]{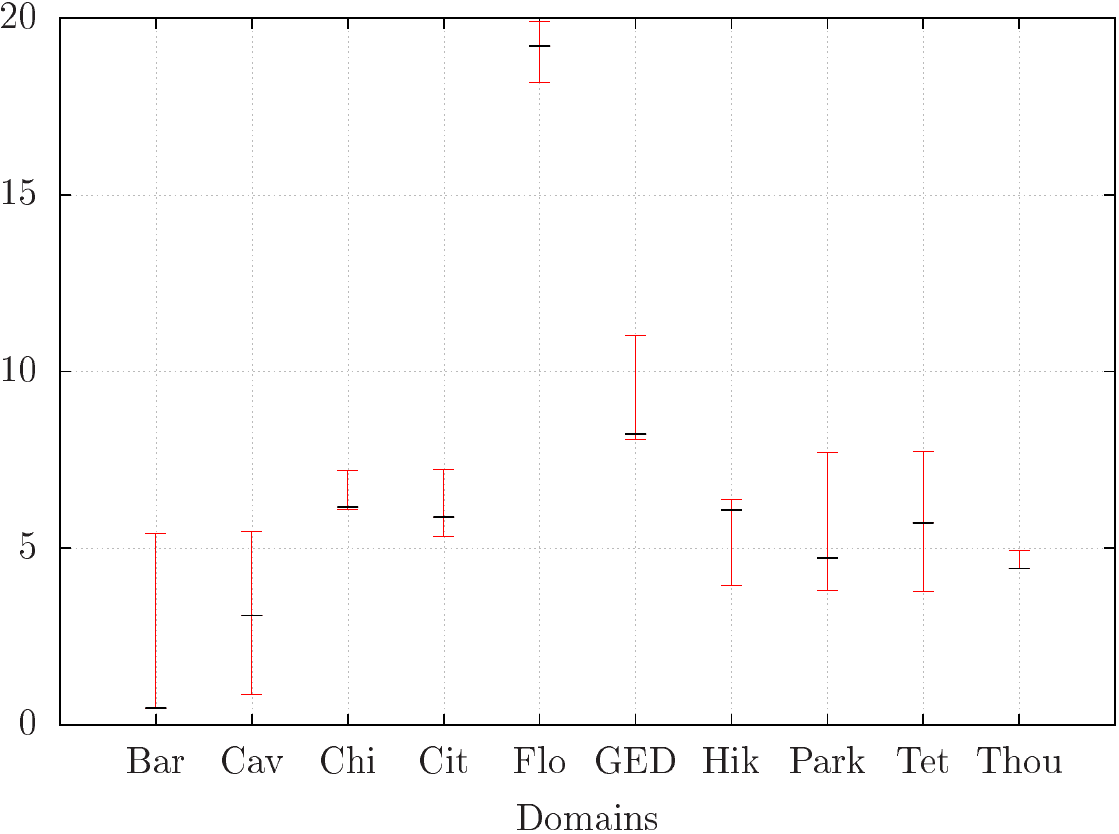}
    \caption{Variation of IPC score observed when running Mpc on benchmarks from the selected domains using heuristically-configured models (red whiskers) and the original domain model (black dash). Shortened Domains' name are reported on x-axis. Transport domain is not shown as Mpc does not solve any instance.}
    \label{fig:depots}
\end{figure}

In order to shed some light on the performance variation of planners running on the heuristically-configured models, Figure \ref{fig:depots} shows maximum and minimum IPC score obtained by Mpc on the selected domains. The performance achieved on the original domain model is indicated by the black dash. Only in the Hiking domain the IPC score obtained on the original domain model is close to the IPC score achieved on the best configured model. In the rest of the cases, the original domain model shows mediocre performance. 
Figure \ref{fig:depots} also shows that in some domains, i.e. Barman, Cave-diving, Parking, and Tetris, the performance gap between the best and the worst configurations is very large: up to 5 IPC score points out of a maximum of 20. In these domains, the proposed heuristics have an exceptional --either positive or negative-- impact on performance, and are therefore able to extract useful information from the PDDL encoding of operators. In the other domain models, we observe that the limited --thus still noticeable-- improvement is due to a number of circumstances, such as the fact that all the operators have a very similar structure, or that there are no actions that are used significantly more (less) often than others.


\begin{table}[!t]


  \begin{center}
          \caption{\label{tab:h2} Best and worst heuristics, according to PAR10 and IPC score. Bold indicates statistically better performance, with regards to the original domain model, according to a Wilcoxon signed rank test. ``ORIG'' indicates the exploitation of the original domain models.} 
    \begin{tabular}{l|c c|c c}
 &  \multicolumn{2}{c|}{Best} & \multicolumn{2}{c}{Worst} \\
 Planner &  PAR10 & IPC & PAR10 & IPC \\
\hline
Bfs-f &  PRE1 & EFF1 & ORIG & PAR1 \\ 
Mpc & 
\textbf{RAT2} & \textbf{RAT2} & RAT1 & RAT1 \\ 
Yahsp3 & 
\textbf{RAT2} & \textbf{RAT1} & PRE2 & PRE2 \\ 
Probe &  PRE1 & PRE1 & ORIG & PAR1 \\ 
arvandherd &  EFF2 & EFF2 & PAR2 & NEG1 \\ 
Freelunch &  EFF2 & EFF2 & ORIG & ORIG \\ 
Mercury & 
\textbf{PRE2} & \textbf{PRE2} & RAT2 & RAT2 \\ 
Jasper &  NEG1 & NEG1 & PAR2 & PAR2 \\ 
SIW & PAR1 & PAR1 & NEG1 & NEG1 \\ 
use & PAR2 & PAR2 & NEG2 & NEG2 \\

\hline
    \end{tabular}
  \end{center}
 
\end{table}

Table~\ref{tab:h1} presents results of the best domain model configurations selected on a domain-by-domain basis. Evidently, in the typical domain-independent scenario, the best configuration is unknown, and a promising heuristic should be picked beforehand and used on every domain. We observed that it is possible to identify a single domain model configuration heuristic for each planner that generally improves its performance across the considered domains. 
Table~\ref{tab:h2} shows the best and the worst domain configuration heuristic in terms of PAR10 and IPC score for each of the considered planners, by considering cumulative results among all the domains. Remarkably, Yahsp3 shows a very specific behaviour with regards to best orderings. Best IPC score is achieved when operators are ordered increasingly, according to the ratio between effects and preconditions; PAR10 is maximised when the inverse ordering is used. Our intuition is that, due to some aspects of the way in which the planner is implemented, the ratio between effects and preconditions is in itself an important factor of the search, as it seems to improve the look-ahead step of the search process. The actual order followed (increasing or decreasing) for listing operators does not seem to be extremely important. Instead, considering only the number of preconditions only, can significantly reduce performance of Yahsp3.  

\begin{figure}[t]
  \centering
    \includegraphics[scale=0.7]{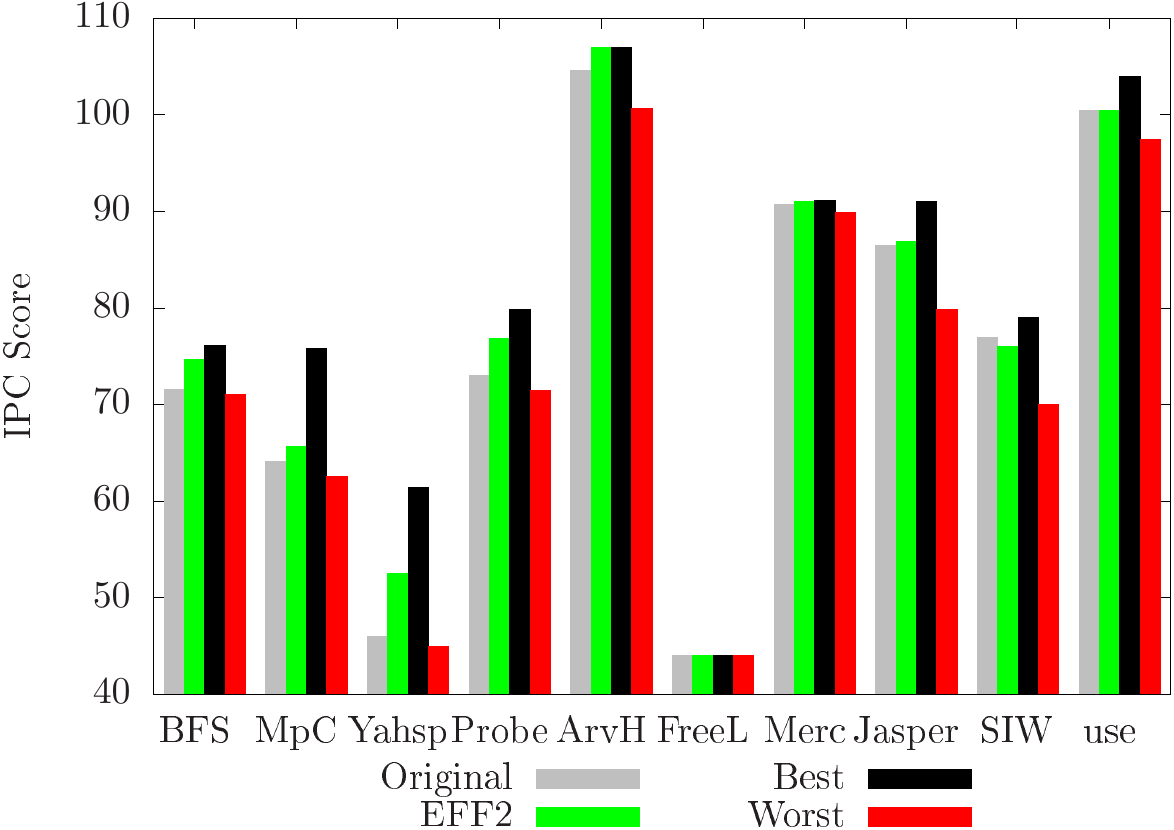}
    \caption{IPC score achieved by planners running on models configured using the single Best heuristic listed in Table \ref{tab:h2} (black), on models resulting by the application of the single Worst heuristic (red), on models configured using the EFF2 heuristic (green), and on the original models (grey). Shortened Planners' name are reported on x-axis. 
    }
    \label{fig:heur}
\end{figure}

For some of the considered planners, the use of the original models (ORIG, in Table~\ref{tab:h2}) leads to the worst possible performance. It has also been observed that, in few cases, performance of planners running on original models is close to the worst possible heuristically-configured models. Cumulative IPC score results are presented in Figure \ref{fig:heur}. 

Interestingly, for four of the planners (Bfs-f, Mpc, Yahsp3, and Jasper) it is possible to identify a domain configuration heuristic that achieves the best runtime performance on most of the domains. For these planners --as shown in Figure \ref{fig:heur}-- results achieved by exploiting the best heuristic selected on a domain-by-domain basis, are very similar to those achieved by considering domain models configured by the single Best heuristic (shown in Table~\ref{tab:h2}): the IPC score performance are reduced by at most 3 points. 
On the other hand, Probe and arvandherd show larger IPC score drops: 15 points the former, 10 the latter. 
The fact that for Probe and arvandherd the best configuration heuristics varies between particular domains suggests that these planners are quite sensitive to the used domain model configuration.


For some planners, using one single planner-specific heuristic for configuring all the domain models (as listed in Table~\ref{tab:h2}) often leads to significantly better performance than the original models (see Figure~\ref{fig:heur}). An example is arvandherd, which performs about 10\% better on the models configured by the single best heuristic than on the original models.

It is not possible to identify the single Best heuristic for each planner, however, 
without the execution of a potentially expensive training session. Unsurprisingly, each planner reacts differently to different domain model configurations and thus there is no ``rule them all'' configuration heuristics. On the other hand, our experimental analysis revealed that the EFF2 heuristic can be a good compromise and outperforms the original models for all the considered planners: \emph{for most of the considered planners it was better to order the operators by increasing number of effects}. The only exception is SIW, for which the EFF2 ordering can have a slightly detrimental impact on performance, when compared to the use of the original domain model. However, the performance achieved by SIW on models configured with the EFF2 ordering are significantly better than those achieved when the worst possible ordering is used, NEG1. 
The differences in the total IPC score for the original models, the EFF2 heuristic and the best planner-specific heuristic are shown in Figure~\ref{fig:heur}. These results indicate that if it is not possible to perform training to identify the best planner-specific heuristic, the EFF2 heuristic provides a reasonable alternative that in general leads to a better performance than the original models.
As shown in Table \ref{tab:h2}, the EFF2 heuristic was also never the worst, but was the best for arvandherd and Freelunch. 

By taking into account the generated plans, we also analysed the impact of our heuristic domain configuration on the quality of plans. Here quality is measured as the cost of the execution of the generated plans, and the analysis has been performed by comparing plans on instances solved by planners run on all the possible models.  For some of the planners, i.e. Bfs-f, Mpc, Freelunch, Mercury, and Jasper, all the plans generated for solving a given problem instance have exactly the same quality. For the remaining planners, plans may be of slightly different quality, but differences are smaller than $5\%$.
Generally, from a quality perspective, for all planners using the models configured according to the proposed heuristics has a \textit{negligible} impact on the structure of generated plans.Overall quality, measured using the IPC quality score, is however increased due to the higher coverage performance given by the use of configured models. 

Figure \ref{fig:heur} gives an overview of how the performance of the considered planners, in terms of IPC score, vary when run using: the original domain model, the models configured using the EFF2 heuristic, the models configured using the best (worst) heuristic as identified in Table \ref{tab:h2}.

\section{Determining the Efficient Positioning of Macros within Domain Models}

Continuing on with the theme of domain model configuration, in this section we
study the question of where to position learned macros in an existing domain model in order to unleash its ``power''.
Macro-operators (macros) are a well known technique
for reformulating planning problems to improve the performance
of planning engines. The optimal ordering of the resulting reformulated
problem, however, has never been studied. This ordering
determines how macros and primitive operators will be used 
together in plan generation, and can make a significant
difference to the utility of the reformulation. For example,
common practice for macro generation techniques is to put newly-generated
macros at the end of the domain model, i.e., after the
primitive operators. In the light of the results presented in the previous sections, which highlight the importance of domain model configuration, it is natural to wonder what impact the placement of the macro in the domain model has, and whether the common practice can be improved upon. 
We study these questions in this section.




\subsection{Methodology and Settings}

We formulate and test a set of hypotheses covering whether the position of
a new macro with respect to primitive operators is significant, whether the normal practice of placing a macro at the end of the
primitive operator set is optimal, and what general insights can be extracted regarding the
relative ordering of macros and operators, so that there is no need to perform highly-expensive domain model configuration.

The planners are chosen based on their performance in the IPCs, their performance in previous studies about macros and reformulation impact (see, e.g., \cite{ma,mum,gsv14}),  
and the range of different techniques they embody: Jasper, LAMA, 
Lpg, Madagascar (Mpc), Probe, SGPlan \cite{sgplan}, and Yahsp3. This has been done in order to consider a pool of planners that have generally good performance, and have shown to be able to effectively exploit macros. 

We focused our study on domains that have been used in learning tracks of IPCs, for which a randomised problem generator is available, and where existing techniques are able to extract generally useful macros for the considered planners. 
Selected domains are: Barman, Blocksworld (4 ops version), Depots, Matching-Bw, Rovers, Satellite, and TPP. We generated 60 testing problems per domain.

We focus most of our study to adding a single macro for each domain, because in most of the domains where existing techniques are able to extract macros, the use of more than one macro has a detrimental effect on planning performance.\footnote{This is due to their grounding, as macros tend to have many parameters derived from the encapsulated operators, and add to the increased branching factor of the search space.} Therefore, focusing on the single macro case allows us to isolate the actual impact of macro position in the domain model, and to cover the vast majority of cases where macro exploitation is fruitful. 

For each domain, we decide which macro to add as follows. First, we extracted several macros using available techniques, namely, Wizard \cite{ma}, MUM \cite{mum}, and MacroFF \cite{macroff} on a set of training instances. As planning engines exploit different planning approaches, the same macro can have a very different impact on each of them; in fact it can be useful for some planners, but it can have a detrimental impact on the performance of others. Among the extracted macros, for each domain we selected the one that generally enables the chosen planners to most improve their performance on the training instances (where the macro was added as the last operator), with respect to the original domain model.
Specifically, whenever possible, we chose the macro that improved the performance of the majority of planners, without having a too strong detrimental effect on the other considered planning systems. In that, we would like to emphasise that this analysis is not about the usefulness of macros, but on selecting how to position macros in a domain model in order to maximise its impact on performance. 

According to these criteria, the following macros were selected:


\begin{itemize}
\item \textbf{Barman}, \texttt{clean-shot(A,B,C,D)}, \texttt{fill-shot(A,E,C,D,F)};
\item \textbf{Blocksworld}, \texttt{unstack(A,B)}, \texttt{putdown(A)};
\item \textbf{Depots}, \texttt{unload(A,B,C,D)},\texttt{drop(A,B,E,D)};
\item \textbf{Matching-BW}, \texttt{putdown-neg-neg(A,B)}, \texttt{pickup(A,C)}, \texttt{stack-neg-neg(A,C,B)};
\item \textbf{Rovers}, \texttt{calibrate(A,B,C,D)}, \texttt{take-image(A,D,E,B,F)};
\item \textbf{Satellite}, \texttt{calibrate(A,B,C)}, \texttt{turn-to(A,D,C)},
\texttt{take-image(A,D,B,E)}, \texttt{turn-to(A,C,D)};
\item  \textbf{TPP}, \texttt{buy(A,B,C,D,E,F,G)}, \texttt{load(A,B,G,F,E,D)}.
\end{itemize}

Within a macro, same parameters' name indicate same objects. 
The macros selected encapsulate 2, 3, or 4 primitive operators.

Hence, domain models are extended (reformulated) by adding the corresponding selected macro. 
For measuring the importance of the position of the macro in the extended domain model, thus synthesising some useful guidelines for macro exploitation and positioning, we considered all the possible position in which the macro can be listed. I.e., starting with an original domain model with $n$ operators, we generated $n+1$ extended domain model configurations, where the macro was listed in $1$st to $(n+1)$-st position. Relative positions of original operators are unchanged. Also, ordering of other elements (i.e., predicates, preconditions and effects) remains the same.


Performance is measured in terms of IPC score, PAR10 and coverage. 
A runtime cutoff of 900 CPU seconds (15 minutes, as in learning tracks of IPC) was used. All the experiments were run on a quad-core 3.0 Ghz CPU, with 4GB of available RAM. In order to account for randomised algorithms and noise, results provided are averaged across three runs. Where possible, seeds of planners have been fixed. 

\begin{table}[t]

\footnotesize

\centering
        \caption{\label{tab:h1adetails} Domain by domain results, in terms of PAR10, coverage (\%), and IPC scores of planners running on domain models where the added macro is in the Best (B) or Worst (W) possible position, according to PAR10, or a the end of the model (L). Bold indicates statistically different PAR10 performance.
        IPC score was calculated separately for each planner; 
        therefore, it cannot be compared across planners. (Partial table, continued in Table \ref{tab:h1bdetails})
}
    \begin{tabular}{l|ccc|ccc|ccc}
    & \multicolumn{3}{|c|}{PAR10} & \multicolumn{3}{|c|}{Coverage} & \multicolumn{3}{c}{IPC score} \\
   
 Planner & B & L & W & B & L &W & B & L & W\\
\hline
\multicolumn{10}{c}{\bf Barman}  \\
\hline
Jasper & 3850.6 & 4150.5 & 4289.9 & 53.3 & 50.3 & 48.3 & 31.7 & 29.0 & 28.6 \\ 
LAMA & 4911.6 & 4911.7 & 4911.7 & 41.7  & 41.7 & 41.7 & 25.0 & 24.5 & 24.5\\ 
Lpg & -- & -- & -- & 0.0 & 0.0 & 0.0 & 0.0 & 0.0 & 0.0 \\ 
Mpc & 4920.1 & 4920.2 & 4920.2 & 50.0 & 50.0 & 50.0 & 23.7 & 23.4 & 23.3 \\ 
Probe & \textbf{4151.0} & 4187.3 & 4762.0 & 50.0 & 50.0  & 43.3 & 28.8 & 28.5 & 24.4 \\ 
SGPlan & 110.8 & 111.4 & 111.4 & 100.0 & 100.0 & 100.0 & 59.8 & 59.0 & 59.0 \\ 
Yahsp3 & 7245.4  & 7245.4 & 7246.0 & 18.3  & 18.3 & 18.3 & 11.0 & 11.0 & 10.9 \\ 
\hline
\multicolumn{10}{c}{\bf BlocksWorld}  \\
\hline
Jasper & \textbf{2783.1} & 3931.5 & 3931.5 & 71.7 & 58.3 & 58.3 & 42.4 & 33.2 & 33.2 \\ 
LAMA & \textbf{1190.7} & 1509.4 & 1509.4 & 90.0 & 86.7 & 86.7 & 52.6 & 48.4 & 48.4 \\ 
Lpg & \textbf{235.9} & 246.8 & 246.8 & 100.0 & 100.0 & 100.0 & 58.0 & 56.9 & 56.9 \\ 
Mpc & \textbf{271.1} & 342.2 & 423.8 & 100.0 & 100.0 & 100.0 & 60.0 & 55.2 & 50.3 \\ 
Probe & \textbf{371.5} & 385.2 & 390.7 & 100.0 & 100.0 & 100.0 & 60.0 & 59.0 & 58.8 \\ 
SGPlan & -- & --  & -- & 0.0 & 0.0  & 0.0 & 0.0& 0.0 & 0.0 \\ 
Yahsp3 & \textbf{0.8} & 1003.9 & 1003.9 & 100.0 & 90.0 & 90.0 & 60.0 & 21.6 & 21.6 \\ 
\hline\multicolumn{10}{c}{\bf Depots}  \\
\hline
Jasper & 5610.6  & 5748.7 & 5748.7 & 38.3 & 36.7 & 36.7 & 22.3 & 21.8 & 21.8 \\ 
LAMA & 5390.4 & 5390.9 & 5390.9 & 41.7 & 41.7 & 41.7 & 24.9  & 24.9 & 24.9 \\ 
Lpg & \textbf{2438.4 }& 3017.3& 3197.2 & 73.3 & 66.7 & 65.0 & 42.1 & 36.1 &  34.2 \\ 
Mpc & 3184.0 & 3187.6 & 3191.3 & 65.0 & 65.0 & 65.0 & 36.8 & 36.5 & 36.0 \\ 
Probe & 38.7 & 39.6 & 40.6 & 100.0 & 100.0 & 100.0 & 58.6 & 57.8  & 57.3 \\ 
SGPlan & 4234.6 & 4291.6 & 4542.8 & 53.3 & 53.3 & 50.0 & 31.2 & 30.1 & 28.4 \\ 
Yahsp3 & 1102.1 & 1117.1 & 1411.1 & 88.3 & 88.3 & 85.0 & 48.2 & 47.9 & 44.3 \\ 
\hline\multicolumn{10}{c}{\bf Matching-BW}  \\
\hline
Jasper & \textbf{2018.6} & \textbf{2018.6} & 2920.6 & 78.3 & 78.3 & 68.3 & 46.5  & 46.5  & 37.5 \\ 
LAMA & 194.6 & 234.6 & 234.6 & 100.0 & 100.0 & 100.0 & 58.8 & 57.3 & 57.3 \\ 
Lpg & 3079.4 & 3901.3 & 3934.4 & 66.7 & 56.7 & 56.7 & 33.5 & 29.9 & 29.6 \\ 
Mpc & 8713.3 & 8821.8 & 8861.1 & 3.3 & 1.7 & 1.7 & 2.0 & 0.9 & 0.6 \\ 
Probe & 2467.0 & 2936.3 & 2936.3 & 73.3 & 68.3 & 68.3 & 41.6 & 38.6 & 38.0 \\ 
SGPlan & 8555.7  & 8851.8 & 8852.6 & 5.0 & 1.7 & 1.7 & 3.0 & 0.6 & 0.5 \\ 
Yahsp3 & 1569.1 & 1599.8 & 2139.2 & 83.3 & 83.3 & 76.7 & 45.3 & 44.3 & 38.6 \\ 
\hline
    \end{tabular}

  
\end{table}

\begin{table}[t]

\footnotesize

\centering
        \caption{\label{tab:h1bdetails} Continuation of Table \ref{tab:h1adetails}. }
    \begin{tabular}{l|ccc|ccc|ccc}
    & \multicolumn{3}{|c|}{PAR10} & \multicolumn{3}{|c|}{Coverage} & \multicolumn{3}{c}{IPC score} \\
   
 Planner & B & L & W & B & L &W & B & L & W\\
\hline\multicolumn{10}{c}{\bf Rovers}  \\
\hline
Jasper & 519.5 & 519.5 & 663.9 & 96.7 & 96.7 & 95.0 & 55.8 & 55.8 & 54.4 \\ 
LAMA & 523.1 & 665.3 & 665.3 & 96.7 & 95.0 & 95.0 & 55.8  & 54.4  & 54.4 \\ 
Lpg & 48.7 & 54.5 & 195.1 & 100.0 & 100.0  & 98.3 & 57.3 & 56.4 & 56.0 \\ 
Mpc & 3956.6 & 3960.3 & 3967.0 & 55.0 & 55.0 & 55.0 & 32.9 & 32.8 & 32.8 \\ 
Probe & 846.9 & 846.9 & 1278.2 & 90.0 & 90.0 & 85.0 & 53.0 & 53.0 & 49.9 \\ 
SGPlan & 62.2 & 63.2 & 63.2 & 100.0 & 100.0 & 100.0 & 57.9 & 57.3 & 57.3 \\ 
Yahsp3 & 10.7 & 10.7 & 10.7 & 100.0 & 100.0 & 100.0 & 58.0 & 57.8 & 57.8 \\ 
\hline\multicolumn{10}{c}{\bf Satellite}  \\
\hline
Jasper & \textbf{322.1}  & 323.8 & 324.3 & 100.0 & 100.0 & 100.0 & 60.0 & 59.9 & 59.8 \\ 
LAMA & \textbf{283.2} & 286.3 & 286.3 & 100.0 & 100.0 & 100.0 & 60.0 & 59.7 & 59.7 \\ 
Lpg & \textbf{8.8} & 9.3 & 10.8 & 100.0 & 100.0 & 100.0 & 59.2 & 56.3 & 54.5 \\ 
Mpc & \textbf{34.6} & 36.9 & 37.8 & 100.0 & 100.0 & 100.0 & 60.0 & 58.1 & 57.7 \\ 
Probe & 662.1 & 681.1 & 1058.6 & 98.3 & 98.3 & 93.3 & 56.9 & 56.0 & 54.6 \\ 
SGPlan & \textbf{9.3} & \textbf{9.3} & 9.4 & 100.0 & 100.0 & 100.0 & 60.0 & 60.0 & 59.8 \\ 
Yahsp3 & \textbf{1.8} & \textbf{1.8} & 1.9 & 100.0 & 100.0 & 100.0 & 60.0 & 60.0 & 59.7 \\ 
\hline\multicolumn{10}{c}{\bf TPP}  \\
\hline
Jasper & 1621.9 & 1621.9 & 1622.9 & 83.3 & 83.3 & 83.3 & 49.9 & 49.9 & 49.8 \\ 
LAMA & 1782.9 & 1782.9 & 1784.6 & 81.7 & 81.7 & 81.7 & 48.9 & 48.9 & 48.8 \\ 
Lpg & 4530.2 & 4802.4 & 4810.4 & 50.0 & 46.7 & 46.7 & 27.1 & 23.9 & 23.7 \\ 
Mpc & 1701.9 & 1986.1 & 1986.1 & 81.7 & 78.3 & 78.3  & 45.2 & 44.0 & 44.0 \\ 
Probe & 127.7 & 128.8 & 129.2 & 100.0 & 100.0 & 100.0 & 59.5 & 59.5  & 59.5 \\ 
SGPlan & 250.1 & 387.6 & 387.6 & 98.3 & 96.7 & 96.7 & 58.7 & 56.5  & 56.5 \\ 
Yahsp3 & 3.4 & 3.4 & 3.4 & 100.0 & 100.0 & 100.0 & 59.9 & 59.8 & 59.8 \\ 
\hline
    \end{tabular}

  
\end{table}

\subsection{Experimental Results}

For the sake of readability and clarity, the experimental analysis is designed around five main hypotheses.

\begin{observation}
The position where macros are listed in the extended domain model affects the runtime performance of considered planners.
\end{observation}


Domain-by-domain performance of the planners is presented in Tables \ref{tab:h1adetails} and \ref{tab:h1bdetails}. In two domains (Blocksworld and Satellite), the position of macros has a statistically significant impact on performance of almost all the considered planners. In several other domains (Barman, Depots, and Matching-BW), all the planners are still affected by the position of the included macro, but only a few show statistically significant gaps. Finally, in two domains (Rovers and TPP) the impact of the macro's position tends to be limited, even though still noticeable for some of the planners (e.g., Lpg and Probe). Also, in Rovers and TPP, the exploitation of macros had a limited impact on performance among the selected planners. Therefore, our empirical evidence suggests that {\it the position of macros in the extended domain model significantly affects planners' performance in cases where macros exploitation is useful}.
In cases where beneficial macro use is limited, we cannot draw the same conclusion. Results shown in Tables \ref{tab:h1adetails}, \ref{tab:h1bdetails} also indicate that the impact of macros' position is not directly related to the number of encapsulated operators: statistically significant differences can also be observed in domains where selected macros include three operators (Matching-BW) and four operators (Satellite). 
By analysing generated solutions, we observed that the impact of macros on runtime performance is not automatically reflected in a larger number of occurrences of macros in plans. That is because macros might influence planning techniques in various ways, e.g., by escaping from local heuristic minima. 

\begin{figure}[t]
  \centering
    \includegraphics[scale=0.6]{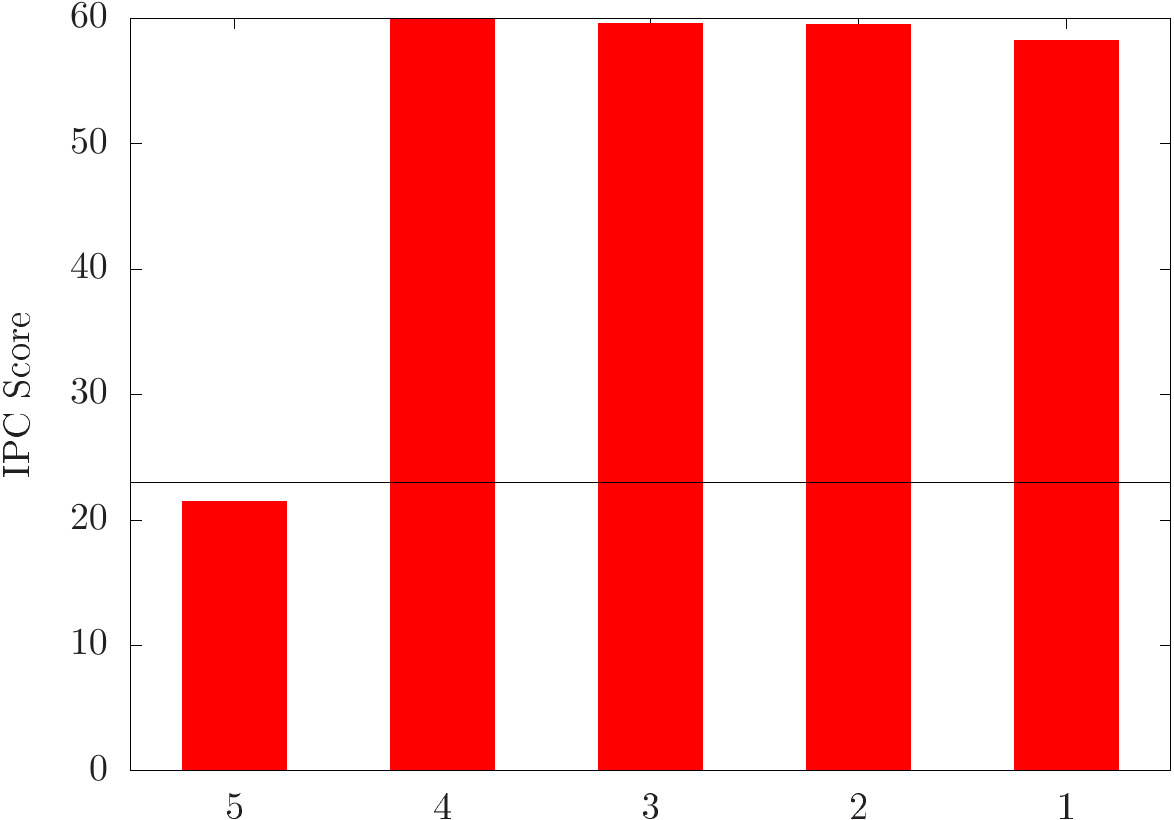}
    \caption{Performance, in terms of IPC score, achieved by Yahsp3 on Blocksworld benchmarks, according to the position of the macro in the domain model. 5 (1) indicates that the macro operator is the last (first) listed operator. The line denotes the IPC score achieved by Yahsp3 on the original domain model (i.e., not extended with a macro).\label{fig:yahsp2}}
\end{figure}

In two cases, it has been observed that the position of the macro can also determine whether its inclusion in the domain model is beneficial or detrimental for the performance of the planner. Lpg with the original Depots domain model achieves better performance ($+3.3\%$ coverage, $+1.8$ IPC score, and $-205.0$ PAR10) than the extended model with the macro inserted in the position that lead to worst possible results, labelled as W in Table \ref{tab:h1g}. Similar behaviour has been observed in Blocksworld for Yahsp3 and is graphically presented in Figure \ref{fig:yahsp2}, where the black line represents the IPC score of Yahsp3 with the original domain model. Also PAR10 is in favour of the original domain model ($-5.2$).

\begin{observation}
Adding macros at the end of the domain model leads to best runtime performance for considered planners.
\end{observation}

It is common practice for macro extracting techniques to extend the original domain model by adding newly-generated macros at the end of the model. This makes the process easier for the extraction tool, but the impact of this ordering choice on the performance of planners has been unclear up to now. Interestingly, our empirical results suggest that common practice does not lead to best runtime results, and this hypothesis is therefore not confirmed. 

\begin{table}[t]


\centering
        \caption{\label{tab:h1g} Results in terms of PAR10, percentages of solved problems and IPC scores of planners running on domain models where the added macro is in the Best or Worst possible position, according to PAR10. All the benchmark domains are considered. Bold indicates statistically different performance in terms of IPC score and PAR10. IPC score was calculated separately for each planner; therefore, it cannot be compared across planners.}
    \begin{tabular}{l|cc|cc|cc}
    & \multicolumn{2}{|c|}{PAR10} & \multicolumn{2}{|c|}{Solved (\%)} & \multicolumn{2}{c}{IPC score} \\
   
 Planner & B & W & B & W & B & W\\
\hline
Jasper & {\bf 2402.1} & 2777.8 & 75.1 & 70.7 & 307.0 & 285.5 \\
LAMA & 2012.1 & 2084.3 & 79.7 & 79.2 & 326.0 & 318.0 \\
Lpg & {\bf 1438.8} & 1940.5 & 73.8 & 68.8 & 277.2 & 254.1 \\ 
Mpc & {\bf 2796.4} & 2887.2 & 64.7 & 64.7 & 256.2 & 247.1 \\
Probe & {\bf 1162.1} & 1417.9 & 88.4 & 86.0 & 358.4 & 342.8 \\ 
SGPlan & {\bf 2245.7} & 2372.0 & 64.5 & 63.3 & 265.6 & 256.4 \\ 
Yahsp3 & {\bf 1355.3} & 1628.9 & 85.0 & 82.1 & 342.3 & 292.7 \\
\hline
\end{tabular}

\end{table}

Table \ref{tab:h2m} shows for each planner on each considered domain, whether exploiting the extended domain model where the macro has been attached at the end leads to best PAR10 performance, or not. It is apparent that in most of the cases, common practice does not allow planning engines to fully exploit the knowledge provided under the form of macros. However, results presented in Table \ref{tab:h2m} indicate some variability among the considered planners: Jasper, for instance, achieves best performance in 3 domains by exploiting the model extended with the macro placed at the end; on the other hand, on 2 other domains this choice led to the worst performance for Jasper.
Other planning engines, such as Lpg and Probe, show less extreme variations: for these, adding macros at the end of models neither led to the worst nor the best performance. 

Moreover, Table \ref{tab:h2m} also shows the position of the macro operator that allows each planner to achieve the best PAR10 performance on the considered domain model. In Blocksworld, Depots, and Satellite, the position of macro is usually the same --or very similar-- among all the planners. Remarkably, they are the domains where the exploitation of macros has a significant impact on planners' performance.

Table \ref{tab:h1g} shows the cumulative performance gap, among all the considered domains, of planners running on the extended domain model with macros placed in the best (B) and the worst (W) position, according to PAR10. Remarkably, for all considered planners but LAMA, the performance difference is statistically significant. The results presented in Table \ref{tab:h1g} also show that coverage can be affected to some extent, with differences up to $5\%$. 
These results indicate that the position of macros in the extended domain model generally has a substantial impact on the planners' performance.
We emphasise that --in contrast to the results from Sections \ref{sec:impact} and \ref{sec:configuration}, where we considered reordering all operators-- the results reported here are solely due to moving the macro operator to a different position in the domain model. The fact that the effects are nevertheless often substantial underlines the importance of this seemingly minor design decision.

\begin{observation}
Performance can vary significantly depending on the relative ordering of a macro and the primitive operators the macro encapsulates.
\end{observation}

In order to verify this hypothesis, we analysed two of the considered domains in detail: Blocksworld and Depots. In Blocksworld, planners are generally very sensitive to the macro's position (see Table \ref{tab:h1adetails}), and macros are extensively used (see Table \ref{tab:h3} for average number of macros per plan).  
The macro used in this domain encapsulates the \texttt{unstack(A,B) + putdown(A)} sequence.  Arguably, this macro is expected to be used very often in generated plans, for removing blocks from towers, in order to make available the blocks below. From this perspective, the most important operator is the \texttt{unstack} operator, and \texttt{putdown} is usually exploited only for freeing the gripper. 
We observed that when the macro is listed in the model \emph{before} the \texttt{unstack} operator, performance of planners is statistically significantly improved, compared adding the macro at the end of the model. For instance, in the case of Yahsp3, penalised average runtime (PAR10) is reduced by over a factor of 1000: from 1003.9 to 0.8. Similar improvements, up to three orders of magnitude, have been observed also when considering the runtime of the planner.
In contrast, when the macro operator is listed \emph{after} the \texttt{unstack} operator, planners' performance deteriorates. An analysis of the generated plans indicates that the macro is not used as much in this case. 
Remarkably, two planners (Lpg and Jasper) achieve even better performance when the macro is listed before {\it both} the operators it encapsulates. 

\begin{table}[t]


\centering
      \caption{\label{tab:h2m} $\blacksquare$ (F) indicates domains where adding macro at the end (top) of the domain model leads to the best PAR10 performance of the considered planner. B (A) indicates that the best performance is achieved when macro operator is listed before (after) the first encapsulated operator. ``b'' represents that the operator has been positioned between the encapsulated operators. ``--'' indicates that no instances from the domains were solved by the considered planner. Domains' name are shortened. ``Total'' gives the number of domains where the last position is the best for adding a macro. } 
    \begin{tabular}{l|ccccccc|c}
   
 Planner & Bar & Bw & D & Mbw & R & S & T & Total\\
\hline
Jasper & A & B(b) & B(b) & $\blacksquare$ & $\blacksquare$ & F & $\blacksquare$  & 3 \\ 
LAMA & F & B & B(b) & F & b & B & $\blacksquare$  & 1 \\ 
Lpg & -- & F & B(b) & A & A & B & B(b) & 0 \\ 
Mpc  & -- & B(b) & B(b) & b & b & B & F & 0 \\ 
Probe & F & B(b) & B & b & $\blacksquare$  & B & b  & 1 \\ 
SGPlan & A  & -- & B(b) & A & B & $\blacksquare$ & B(b) & 1 \\ 
Yahsp3 & $\blacksquare$ & B(b) & B(b) & A & B & $\blacksquare$ & B(b) & 2 \\ 
\hline
    \end{tabular}
  
\end{table}

Differently from what has been observed in the Blocksworld domain, in Depots the considered macro is rarely used (approximately five macros per plan, and plans include few dozens of actions), and planners tend to be less sensitive to its position in the model. 
The macro used in this domain encapsulates the \texttt{unload(A,B,C,D) + drop(A,B,E,D)} sequence. In the Depots domain the \texttt{drop} operator is listed in the domain model before the other \texttt{unload} operator. Interestingly, as shown in Table \ref{tab:h2m}, planners tend to achieve the best runtime performance when the macro is listed before the first encapsulated operator, but after the second operator: all the planners show their best performance when the macro operator is listed \textit{between} these two operators. 

In order to confirm the above observations, we re-ordered the list of operators of the two domain models, according to the best configuration identified in Section \ref{ijcai15}; the configured models showed to be generally good for all the planners. 
We then extended the configured domain models by listing the macro in all the possible positions, and re-ran the planners. In Blocksworld, we expected that best performances are achieved when the macro is listed as the second operator of the model, i.e., just before the \texttt{unstack} operator. Indeed, this was the case for all planners except Mpc and Jasper, which achieve better performance when the macro is listed as the first operator of the model. Furthermore, it is generally true that the best possible performance is achieved by planners when the macro is listed \emph{after} the first operator it encapsulates (e.g. after \texttt{unstack} in the Blocksworld), and \emph{before} the second operator (\texttt{putdown} in the Blocksworld). 

Results have been also confirmed when considering the configured Depots model: Mpc and Jasper show the best performance when the macro is listed first. On the other hand, all the remaining planners achieve their best performance when the macro is listed between the two encapsulated operators. 

Interestingly, in both the configured models, original operators included in the considered macros swapped their position with regards to the original domain models. This let us believe that, in cases where the best position of a macro is between encapsulated operators, the order in which those operators are listed is not critical.


Our results on the configured domain models confirm that \textit{planners' performance varies significantly depending on the relative ordering of a macro and the primitive operators the macro encapsulates}.

\begin{figure}[t]
  \centering
    \includegraphics[scale=0.45]{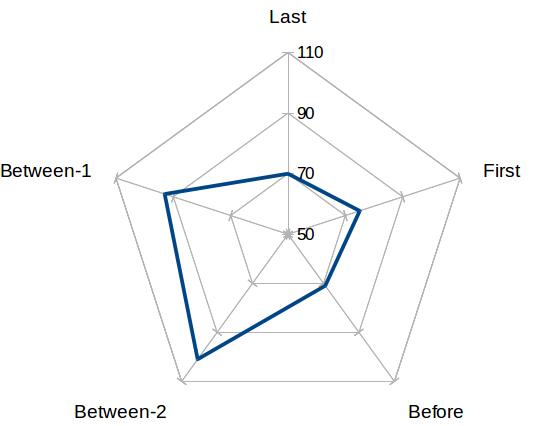}
    \caption{Number of instances on which considered planners achieved fastest performance when exploiting differently extended Depots domain models. ``Between'' (before) indicates that the macro is listed between (before both of) the encapsulated operators. Note that there are two positions between encapsulated operators. ``Last'' (First) indicates that the macro is listed at end (top) of the model.}
    \label{fig:depots2}
\end{figure}

\begin{observation}
It is possible to identify the position of a macro which will maximise the possibility of performance improvement of considered planners without actually running the planners.
\end{observation}

Results shown in Table \ref{tab:h2m}, and the investigation presented above, provide evidence indicating that the best position differs across different planners. However, is it possible to identify a position that allows planners to achieve generally good performance and, at the same time, to avoid positions that lead to the worst performance for some planners?

According to the presented experimental results, the best macro position mainly depends on two aspects: relative position of encapsulated operators, and the expected use of the macro in solution plans. 
When a macro is used frequently, the macro operator should be listed early in the model, and before the first encapsulated operator. Planners are then likely to select the macro more often than the primitive operator. On the contrary, listing the macro after the first encapsulated operator, or at the end of the domain model, leads to worse planning performance.

In domains where macros rarely occur in solution plans, it is better to list them \textit{between} encapsulated operators. 
We observed this behaviour for the Depots domain, where planners tend to perform better when the macro is listed \textit{between} encapsulated operators, and sometimes \textit{before} the first operator. Figure \ref{fig:depots2} synthesises the described behaviour by showing, for each considered extended Depots domain model, the number of times a planner achieved best performance. It should be noted that there are two \textit{between} positions, because another primitive operator is listed between the two operators encapsulated in the macro. 
A behaviour similar to the one described in Depots has been observed also in TPP, where the considered macros represent approximately 5\% of plans' actions. 
Instead, in domains where macros do not have a strong beneficial impact on planners' performance, and are rarely exploited in solution plans, such as Barman and Matching-BW, the best position of the macro is not very clear.
According to results presented in Table \ref{tab:h2m} some planners, like Jasper, SGPlan and Yahsp3, achieve best performance when the macro is listed \emph{after} the first encapsulated operator. Probe and LAMA, instead, obtain better performance when the macro is listed \emph{before} the first encapsulated operator. 

\begin{table}[t]


\centering
        \caption{\label{tab:h1derived} Depots results, in terms of PAR10, coverage (\%), and IPC scores of additional IPC 2014 Agile track
planners running on domain models where the added macro is in the Best (B) possible position, according to the results observed in our analysis, or a the end of the model (L), following the common practice. IPC score was calculated separately for each planner; therefore,
it cannot be compared across planners.}
    \begin{tabular}{l|cc|cc|cc}
    & \multicolumn{2}{|c|}{PAR10} & \multicolumn{2}{|c|}{Solved (\%)} & \multicolumn{2}{c}{IPC score} \\
   
 Planner & B & L & B & L & B & L\\
\hline
Bfs-f & 2364.9 & 2554.4 & 61.7 & 58.3 & 37.0 & 34.5 \\
Freelunch & 4119.7 & 4200.8 & 30.0 & 30.0 & 17.9 & 17.9 \\
Mercury & 4505.7 & 4506.0 & 25.0 & 25.0 & 15.0 & 14.8 \\
use & 4106.4 & 4304.2 & 31.7 & 28.3 & 19.0 & 16.9 \\
SIW & 1869.9 & 2069.8 & 70.0 & 66.7 & 40.2 & 39.6 \\
\hline
\end{tabular}

\end{table}

For further validating the above observations, we run some additional experiments using all the remaining domain-independent planners, not based on portfolio approaches, from the IPC 2014 Agile track: Bfs-f, Freelunch, Mercury, use, and SIW. We compared the performance they achieved on the considered benchmarks, either running on the domain model extended according to the common practice, or following the position suggested by our empirical analysis. Results indicate that our hypothesis is correct: the suggested position of macro has a generally positive impact on planners' performance, also in terms of coverage. Moreover, performance achieved are on no domain worse than those delivered when exploiting models extended following the common practice. As an example, Table \ref{tab:h1derived} shows the performance achieved by the planners on the Depots domain. 



 
  

\begin{observation}
The position where macros are listed in the extended domain model affects the quality of plans generated by the set of selected planners.
\end{observation}

In domains where the number of occurrences of macros in generated plans is limited, such as Depots, TPP, Matching-BW, and Barman, the quality of plans (here measured as number of actions, averaged across instances solved by all the models) is not affected by the position where the macro is listed in the domain model. Jasper and LAMA generate exactly the same plans. The other planners show a very limited variability; plans are very similar in terms of macro occurrences. Average plans' length is also very similar: average differences are in the order of  
$1$--$2$ actions for plans between few tens to few hundreds of actions. 

Interestingly, also in domains like Satellite and Rovers, where the number of occurrences of macros in plans is significant, the position of macros in the domain model has a very limited impact on the shape of plans. In Satellite, for instance, the maximum difference between plans generated by Lpg --running on differently extended domain models-- for solving the same problem instance is of $10$ actions, where average plan length is of $252.8$. 
Notably, each plan includes a large amount of macro instantiations: on average, more than $80\%$ of the actions in plans are macros. Similar behaviour has been observed in the Rovers domain. The only planner that is able to generate significantly different plans in Rovers is Probe; plans can be up to $40$ actions shorter, according to the position of the macro in the domain model. Interestingly, the number of macro occurrences in plans generated for solving the same problem does not change, but the rover visits waypoints in a different order. This indicates that the position of the macro in the model affects the search space exploration.

\begin{table}[t]


  \begin{center}
      \caption{\label{tab:h3} Results in terms of average quality of plans (standard deviation) and average number of macros used per plan, by planners running on Blocksworld models where the added macro is in the Best or Worst possible position, according to plan quality. 
        Bold indicates statistically different performance. SGPlan not shown: did not solve any instance.} 
    \begin{tabular}{l|cc|cc}
    & \multicolumn{2}{|c|}{Average Plans Q} & \multicolumn{2}{|c}{Average \# M} \\
   
 Planner & B & W & B & W \\ 
\hline
Jasper  & 387.9 (112.9) & 461.3 (182.6) & 65.3 & 112.8 \\ 
LAMA  & {\bf 572.6 (321.6)} & 625.3 (217.6) & 60.3 & 152.9\\ 
Lpg  & 592.9 (81.1) & 613.9 (99.8) & 138.6 & 140.8\\ 
Mpc  & 321.3 (31.6) & 322.3 (32.2) & 75.2 & 75.4\\ 
Probe  & 298.7 (20.9) & 298.8 (20.8) & 71.9 & 71.8 \\ 
Yahsp3  & {\bf 1155.0 (349.2)} & 3103.0 (1425.5) & 276.7 & 0.7\\ 
\hline
    \end{tabular}
  \end{center}
  
\end{table}

In contrast, we noticed that in Blocksworld, plans are usually strongly affected by the position of the macro in the domain model. Only Mpc and Probe generate very similar-quality plans regardless of the position of the macro. There is not a clear intuition behind such behaviour. It may due, in the case of Mpc, to the use of a SAT-based approach that checks the satisfiability of the problem given a certain temporal horizon, which forces plans to be very similar in terms of overall quality, as soon as solutions can be found with the given horizon. Table \ref{tab:h3} gives planner-by-planner details on the Blocksworld domain. The fact that these results for the Blocksworld domain are qualitatively different than those for the  Rovers and Satellite domains begs the question what is the difference between the macros generated in these cases.
Macros extracted for Rovers and Satellite encapsulate operators that must be executed in that sequence on the involved objects, in order to achieve goals. This is not the case in Blocksworld, where the execution of an \texttt{unstack} action could also be followed by an instance of a \texttt{stack} action. In fact, the macro added to the Blocksworld domain model provides a heuristic to the planner: when a block is unstacked, it is safe to put it on the table. This results in a free gripper, which can either be used for moving other blocks, or stacking the same one somewhere else. In this sense, macro exploitation strongly affects the quality of solutions.


Summarising, the empirical analysis performed in this section provides evidence that \textit{support} hypotheses 1, 3, 4, and 5. In contrast, hypothesis 2 is not been verified.



\subsection{Models extended with multiple macros} \label{sec:multiple_macros}
%

While our experimental analysis so far focused on domain models extended by considering a single macro, there are some cases in which extending the domain model with two macros has a positive impact on performance. For example, in the Blocksworld domain, the use of \texttt{unstack + putdown} and \texttt{pickup + stack} can improve performance. This is the only domain, among those considered in our analysis, where the use of more than one macro improved the performance of planning engines when compared with the original domain model. In the other cases a negative impact is observed, mainly because planners run out of memory. The Blocksworld peculiarity is that macros can fully replace the primitive operators, so their use can halve the depth of a solution in the search space. 

For our analysis, we considered six different Blocksworld extended domain models, corresponding to: macros are attached at the end of the domain model (two versions, according to which macro is listed first), macros listed on top of the domain model (two versions), each macro is listed just before the first encapsulated operator, and each macro is listed just after the the first encapsulated operator. Adding macros at the end of the domain model simulates the common practice, while listing them on top of the domain model should intuitively foster their exploitation. Finally, having them just before (after) the first encapsulated operator follows the observations made in the previous section, and should correspond to the best (worst) possible positioning. The empirical evidence confirms our intuition. For all considered planners, the worst performance is achieved when macros are listed at the end of the domain model, or after the first encapsulated operator. On the contrary, best performance is obtained when macros are listed at the top of the domain model, or just before the first encapsulated operators. Impact can be very significant. Moving macros from before to after the first encapsulated operator,
leads LAMA to drop performance by $33.3\%$, in terms of coverage, and by
$31.2$ IPC score points. Similarly, Yahsp3 solves
every instance in less than $2$ CPU-time seconds when macros are listed in positions before
 the first encapsulated operator. Otherwise Yahsp3 does not solve any instance.

\section{Discussion}

Given the results presented in Section \ref{sec:impact}, it is apparent that comparing and ranking planners using a single domain model configuration may lead to a biased evaluation. In this section we suggest an approach that addresses such issue, without increasing the experimental burden.

What is usually done in competitions like the IPC --or in empirical analysis involving the execution and comparison of domain-independent planners-- is to pick up one random sample from the population of domain model's configurations. While the configuration selected may, or may not, have some interesting characteristics for human experts, it can be considered as completely random from the perspective of a planning engine. This is also because, usually, low-level programming details of planners are not taken into account when models are crafted. 

In strictly statistical terms, the empirical evaluation presented in Section \ref{sec:impact} was based on simple random sampling \cite{thompson2012simple}: 50 distinct configurations have been selected from the domain model's configuration population, in such a way that they have the same probability to be selected. As it is apparent, the larger the considered number of distinct configurations used in the evaluation, the better is the understanding of the planners' performance, and the more accurate is the overall ranking. However, due to the organisational and computational burden involved with running planning (or, more generally, AI) competitions, planners cannot be compared using a large set of differently configured domain models for solving all the benchmarks from the considered domain. It is not even feasible, given the limited available time, to require organisers to run the same competition tests multiple times.  

\begin{figure}[t]
\centering
\includegraphics[width=0.5\textwidth, angle=90]{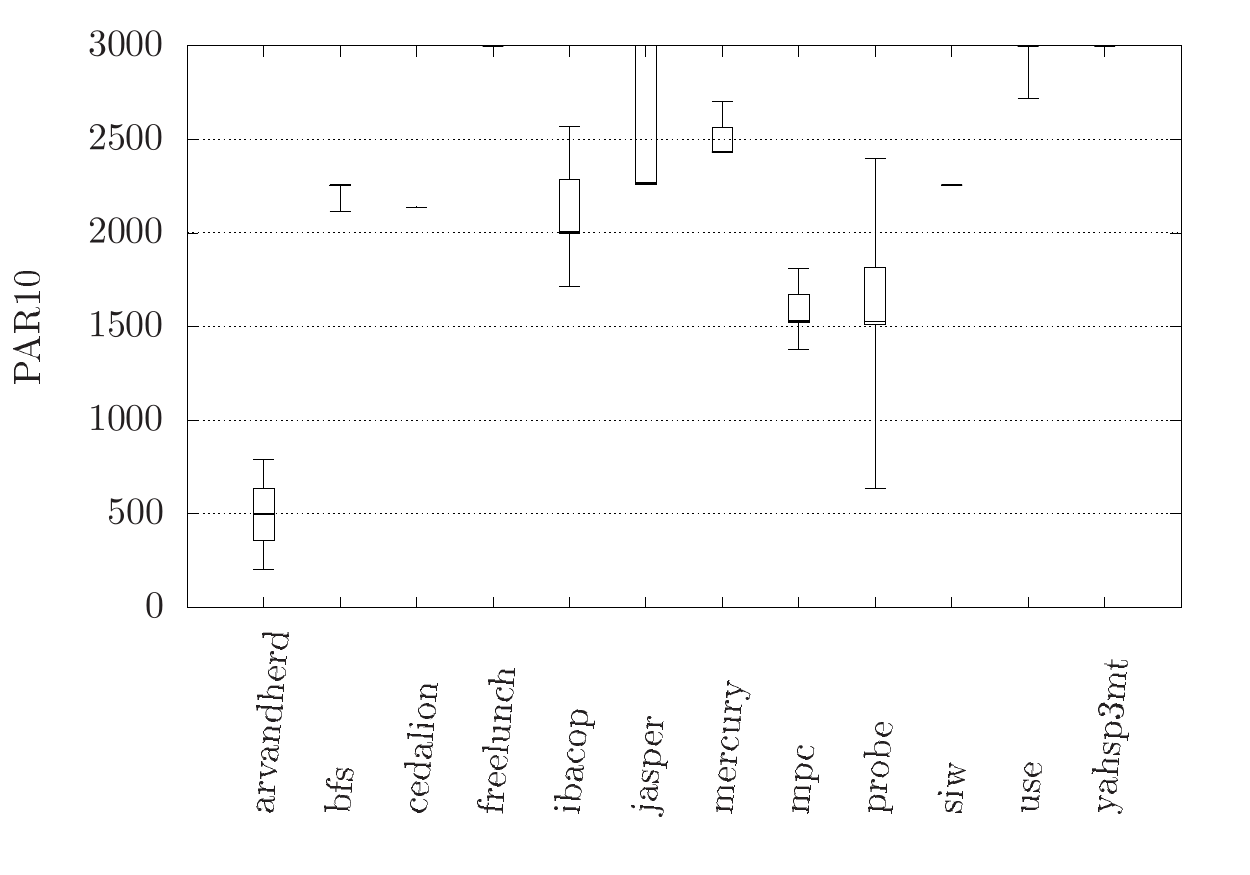}
\centering
\includegraphics[width=0.5\textwidth, angle=90]{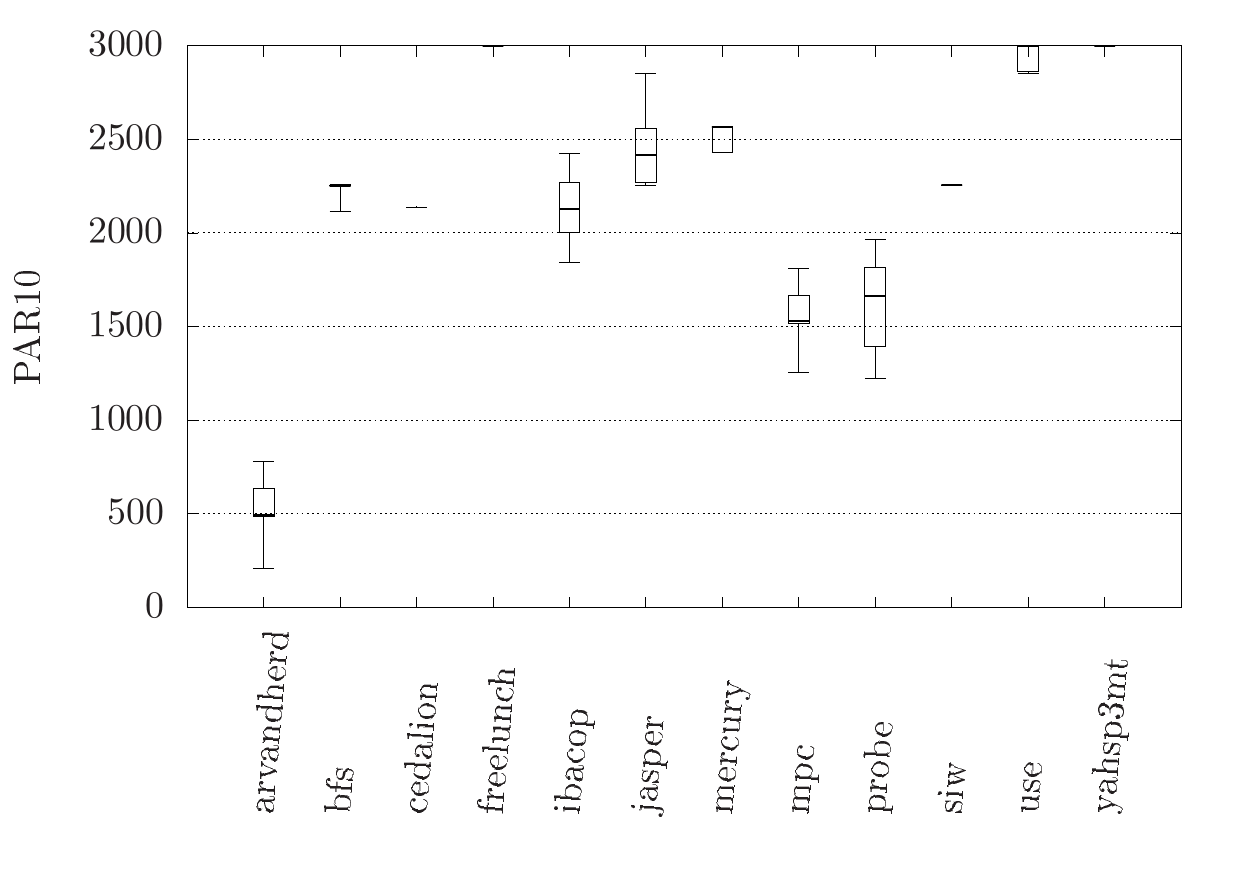}
\caption{PAR10 performance variability of planners when run on a single domain model configuration for solving all 20 CityCar benchmarks (top), and when run on a different configuration per problem instance (Bottom).
Whiskers refer to the best and worst performance achieved by the planner, while the box indicates the first, second, and third quartiles.
\label{fig:randomised}}
\end{figure}

Here we address the problem of getting a more accurate and stable evaluation of planners' performance, without increasing the burden on competition organisers. The idea is to run each problem instance of the benchmark set using a different, and randomly generated, domain model configuration. The underlying assumption is that, by changing the configuration multiple times, we limit the impact on planner's performance of extremely good and extremely bad --from a given planner's perspective-- configurations. A problem-by-problem randomisation should, ideally, allow to provide an overall evaluation of a given planner's performance that is closer to the planner's median, on the considered domain model. Evidently, the additional overhead due to the random configuration of many domain models is negligible, and this ensures a very limited additional burden on competitions' organisers. 

In order to assess the usefulness of the described approach, we focused on one of the domains of the Agile track where a significant variability in planners' performance were observed: CityCar. We drew 100 bootstrap resampling of different domain model configurations for each of the 20 problem instances of the CityCar domain, and we evaluated the PAR10 performance fluctuations of participants. Figure \ref{fig:randomised} shows the results of our analysis, under the form of box and whiskers plots. Whiskers refer to the best and worst performance achieved by the planner, while the box indicates the first, second, and third quartiles. The top graph provides results of the variability of PAR10 performance of planners when run on a single, randomly picked, domain model configuration for solving all the domain benchmarks, while the lower graph reports the fluctuation observed when run a different domain model configuration for each problem instance. PAR10 has been selected, as the PAR10 performance of  given planner are not affected by the performance of the other considered competitors; moreover, PAR10 combines information of runtime and coverage.

The results presented in Figure \ref{fig:randomised} indicate that running a planner on many different domain model configurations for solving benchmarks from the same domain can generally reduce performance fluctuations. This is particularly true for planners like IBaCoP, Probe, use, and Jasper, which can be very sensitive to the models' configuration. In such planners, the impact of outliers (i.e., particularly good or bad configurations) is compensated across benchmarks by using different  configurations. It is worth emphasising that in the considered CityCar domain, Probe can deliver statistically significantly different performance according to the domain model configuration used for solving the benchmarks (see the top part of Figure \ref{fig:randomised}). By considering a different domain model configuration per problem instance, performance discrepancies are dramatically reduced, and the delivered performances are not statistically significantly different any more.
Planners with very limited variations are not negatively affected by the exploitation of the introduced technique. In the light of this, and given the limited additional burden on experiments, we believe that the introduced technique should be routinely employed in empirical evaluations as well as international competitions, for the sake of delivering more reliable and more stable results.



\section{Conclusion}\label{sec:conclusion}

A key product of knowledge engineering for automated planning is the domain model. In this paper, we investigated how the configuration of domain models affects the performance of state-of-the-art
domain-independent planners. We focused on domain models written in PDDL, considering as modifiable the orderings of predicate declarations, the ordering of operators, and the ordering of predicates within preconditions and effects. We first showed the large impact different (random) domain configurations had on performance and then
introduced offline and online configuration techniques for finding better domain configurations in order to improve the performance of given planning engines. Our results indicated that the configuration of domain models can allow to achieve up-to-25-fold speedup. 
Finally, we investigated how the orderings of learned structures (macros), when
added to the original domain model, can affect the performance of planners. On this matter, our empirical analysis highlighted that PAR10 can be improved by up to 3 orders of magnitude by modifying macros' positioning in extended domain models.  

The main empirical finding from this paper is that the ordering of elements of domain
models significantly affects the runtime performance of state of the art domain-independent planners, regardless of the search technique they exploit, or their implementation details.
The configuration of domain models can improve the planning process, often remarkably.
Our thorough experimental analysis provided valuable lessons from which domain engineers can learn how to encode effective and efficient domain models. More specifically:

\begin{itemize}
\item The impact of the configuration of models on planning engines can be limited, in order to obtain a more accurate and stable comparison of planners’ performance, by running each problem instance of the benchmark set using a different, randomly generated, domain model configuration. The intuition is that, by changing the configuration multiple times, we limit the impact of configurations that are extremely good or extremely bad for a given planner and obtain an average over configurations. Furthermore, this approach does not increase the burden on competition organisers.
\item Operators tend to have a very strong impact on performance. In domains with strong directionality (i.e., some actions are very likely to be executed in initial steps and not in final steps, and vice-versa), ordering operators according to this directionality clearly improves planning performance. In domains where all the operators have potentially the same probability
to be executed in any part of the plan, a good operators' order follows the frequency of action usage in the plan: most used first. 
\item Preconditions and effects of operators seem to have a less marked (but still noticeable) impact on the performance of domain-independent planners. Our observations suggest that preconditions of operators which are most likely to be unsatisfied should be listed earlier. On the other hand, the impact of ordering of effect predicates is not very clear; our intuition is that the main effect for which an action is usually executed should be presented earlier.
\item Evidence from our experimental analysis indicates that the configuration of domain models does not significantly affect the quality of generated solution plans.
\item When considering ``extended'' domain models that include macro-operators, our large experimental analysis shows that the position of macros in the extended domain model can have a statistically significant impact on planners' performance. Furthermore, the usefulness of macros for a planner is affected by their positioning in the domain model: a poorly placed macro can have a detrimental impact on performance, while a macro that is carefully listed in the extended model can result in a significant performance boost for the planning engine. The results indicate that if a macro was used heavily in solutions, then listing it first would tend to speed up plan generation. 
\end{itemize}

We see several other avenues for future work. We plan to investigate the impact of domain model configuration on non-classical planning models, such as temporal and mixed discrete-continuous. We are
interested in exploring the configuration of models encoded
in languages different from PDDL, such as SAS+ and SAT.
Such analysis can also shed some light on the different impact
that model configuration has on planning systems that exploit
different techniques. We plan to combine domain model configuration with portfolio approaches, to automatically configure a portfolio of domain models, for one or more planners. We are interested in improving existing macro generation techniques, in order to allow them to take into account macros' positioning, so that potentially useful macros are not discarded only because of their position. We are also interested in exploring domain-independent heuristic orderings that can be incorporated in planners, to reduce the bias due to the way in which models are presented. We envisage the use of domain model configuration as an approach to evaluate the way in which models are engineered: in cases where significant performance improvement can be achieved by re-ordering operators, it may be possible to reformulate the model to uniform gains for all planners. 
Finally, we are considering investigating the interactions of domain model configuration and online configuration of problem models, which could potentially lead to
further performance improvements and to useful insights on the obscure task of describing and modelling problems.




\bibliographystyle{plain}
\bibliography{reference}   

\begin{thebibliography}{10}

\bibitem{ai2004mapgen}
Mitchell Ai-Chang, John Bresina, Len Charest, Adam Chase, JC-J Hsu, Ari
  Jonsson, Bob Kanefsky, Paul Morris, Kanna Rajan, Jeffrey Yglesias, et~al.
\newblock Mapgen: mixed-initiative planning and scheduling for the mars
  exploration rover mission.
\newblock {\em IEEE Intelligent Systems}, 19(1):8--12, 2004.

\bibitem{AnsoteguiST09}
C.~Ans{\'{o}}tegui, M.~Sellmann, and K.~Tierney.
\newblock A gender-based genetic algorithm for the automatic configuration of
  algorithms.
\newblock In {\em Proceedings of the Principles and Practice of Constraint
  Programming ({CP})}, pages 142--157, 2009.

\bibitem{DBLP:conf/aips/ArecesBD014}
Carlos Areces, Facundo Bustos, Mart{\'{\i}}n~Ariel Dominguez, and J{\"{o}}rg
  Hoffmann.
\newblock Optimizing planning domains by automatic action schema splitting.
\newblock In {\em Proceedings of the Twenty-Fourth International Conference on
  Automated Planning and Scheduling, {ICAPS}}, 2014.

\bibitem{freelunch}
T.~Balyo.
\newblock The freelunch planning system entering ipc 2014.
\newblock In {\em Proceedings of the 8th International Planning Competition
  (IPC-2014)}, 2014.

\bibitem{balyo2013relaxing}
Tomas Balyo.
\newblock Relaxing the relaxed exist-step parallel planning semantics.
\newblock In {\em 2013 IEEE 25th International Conference on Tools with
  Artificial Intelligence}, pages 865--871, 2013.

\bibitem{macroff}
A.~Botea, M.~Enzenberger, M.~M{\"u}ller, and J.~Schaeffer.
\newblock Macro-ff: Improving ai planning with automatically learned
  macro-operators.
\newblock {\em Journal of Artificial Intelligence Research (JAIR)},
  24:581--621, 2005.

\bibitem{breiman2001random}
L.~Breiman.
\newblock Random forests.
\newblock {\em Machine learning}, 45(1):5--32, 2001.

\bibitem{DBLP:journals/jair/CenamorRF16}
Isabel Cenamor, Tom{\'{a}}s de~la Rosa, and Fernando Fern{\'{a}}ndez.
\newblock The ibacop planning system: Instance-based configured portfolios.
\newblock {\em J. Artif. Intell. Res.}, 56:657--691, 2016.

\bibitem{ijar17}
Federico Cerutti, Mauro Vallati, and Massimiliano Giacomin.
\newblock On the impact of configuration on abstract argumentation automated
  reasoning.
\newblock {\em International Journal of Approximate Reasoning}, 2017.

\bibitem{my_ker}
L.~Chrpa.
\newblock Generation of macro-operators via investigation of action
  dependencies in plans.
\newblock {\em Knowledge Engineering Review}, 25(3):281--297, 2010.

\bibitem{my_sara}
L.~Chrpa and R.~Bart\'{a}k.
\newblock Reformulating planning problems by eliminating unpromising actions.
\newblock In {\em Symposium on Abstraction, Reformulation, and Approximation,
  SARA}, pages 50--57, 2009.

\bibitem{my_ecai}
L.~Chrpa and T.~L. McCluskey.
\newblock On exploiting structures of classical planning problems: Generalizing
  entanglements.
\newblock In {\em European Conference on Artificial Intelligence, ECAI}, pages
  240--245, 2012.

\bibitem{mum}
L.~Chrpa, M.~Vallati, and T.~L. McCluskey.
\newblock Mum: A technique for maximising the utility of macro-operators by
  constrained generation and use.
\newblock In {\em Proceedings of the International Conference on Automated
  Planning and Scheduling, {ICAPS}}, pages 65--73, 2014.

\bibitem{ma2}
A.~Coles, M.~Fox, and A.~Smith.
\newblock Online identification of useful macro-actions for planning.
\newblock In {\em the International Conference on Automated Planning and
  Scheduling, ICAPS}, pages 97--104, 2007.

\bibitem{fdauto}
C.~Fawcett, M.~Helmert, H.H. Hoos, E.~Karpas, G.~R{\"o}ger, and J.~Seipp.
\newblock Fd-autotune: Domain-specific configuration using fast-downward.
\newblock In {\em {Workshop on Planning and Learning (PAL)}}, 2011.

\bibitem{pddl2.1}
Maria Fox and Derek Long.
\newblock {PDDL2.1:} an extension to {PDDL} for expressing temporal planning
  domains.
\newblock {\em J. Artif. Intell. Res. {(JAIR)}}, 20:61--124, 2003.

\bibitem{lpg}
A.~E. Gerevini, A.~Saetti, and I.~Serina.
\newblock Planning through stochastic local search and temporal action graphs
  in {LPG}.
\newblock {\em Journal of Artificial Intelligence Research (JAIR)},
  20:239--290, 2003.

\bibitem{gsv14}
A.~E. Gerevini, A.~Saetti, and M.~Vallati.
\newblock Planning through automatic portfolio configuration: The pbp approach.
\newblock {\em J. Artif. Intell. Res. {(JAIR)}}, 50:639--696, 2014.

\bibitem{pddl}
M.~Ghallab, C.~Knoblock Isi, S.~Penberthy, D.~E Smith, Y.~Sun, and D.~Weld.
\newblock Pddl - the planning domain definition language.
\newblock Technical report, 1998.

\bibitem{apl}
M.~Ghallab, D.~Nau, and P.~Traverso.
\newblock {\em Automated planning, theory and practice}.
\newblock Morgan Kaufmann Publishers, 2004.

\bibitem{DBLP:journals/jair/Helmert06}
Malte Helmert.
\newblock The fast downward planning system.
\newblock {\em J. Artif. Intell. Res.}, 26:191--246, 2006.

\bibitem{HoweD02}
A.~E. Howe and E.~Dahlman.
\newblock A critical assessment of benchmark comparison in planning.
\newblock {\em J. Artif. Intell. Res. (JAIR)}, 17:1--33, 2002.

\bibitem{sgplan}
Chich-Wie Hsu and Benjamin~W. Wah.
\newblock The sgplan planning system in ipc-6.
\newblock In {\em The 6th International Planning Competition (IPC-6)}, 2008.

\bibitem{smac}
F.~Hutter, H.~H. Hoos, and K.~Leyton-Brown.
\newblock Sequential model-based optimization for general algorithm
  configuration.
\newblock In {\em Proceedings of the 5th Learning and Intelligent OptimizatioN
  Conference ({LION})}, pages 507--523, 2011.

\bibitem{hutter2014efficient}
F.~Hutter, H.~H. Hoos, and K.~Leyton-Brown.
\newblock An efficient approach for assessing hyperparameter importance.
\newblock In {\em Proceedings of The 31st International Conference on Machine
  Learning}, pages 754--762, 2014.

\bibitem{HutterHLS09}
F.~Hutter, H.~H. Hoos, K.~Leyton{-}Brown, and T.~St{\"{u}}tzle.
\newblock Paramils: An automatic algorithm configuration framework.
\newblock {\em J. Artif. Intell. Res. {(JAIR)}}, 36:267--306, 2009.

\bibitem{cssc}
F.~Hutter, M.~Lindauer, A.~Balint, S.~Bayless, H.~Hoos, and K.~Leyton-Brown.
\newblock The configurable sat solver challenge (cssc).
\newblock {\em Artificial Intelligence Journal (AIJ)}, 243:1--25, February
  2017.

\bibitem{epms}
F.~Hutter, L.~Xu, H.~H. Hoos, and K.~Leyton-Brown.
\newblock Algorithm runtime prediction: Methods \& evaluation.
\newblock {\em Artificial Intelligence}, 206:79--111, 2014.

\bibitem{mercury}
M.~Katz and J.~Hoffmann.
\newblock Mercury planner: Pushing the limits of partial delete relaxation.
\newblock In {\em Proceedings of the 8th International Planning Competition
  (IPC-2014)}, 2014.

\bibitem{mo}
R.E. Korf.
\newblock Macro-operators: A weak method for learning.
\newblock {\em Artificial Intelligence}, 26(1):35--77, 1985.

\bibitem{probe}
N.~Lipovetzky, M.~Ramirez, C.~Muise, and H.~Geffner.
\newblock Width and inference based planners: Siw, bfs(f), and probe.
\newblock In {\em Proceedings of the 8th International Planning Competition
  (IPC-2014)}, 2014.

\bibitem{McCPort}
T.~L. McCluskey and J.~M. Porteous.
\newblock Engineering and compiling planning domain models to promote validity
  and efficiency.
\newblock {\em Artificial Intelligence}, 95(1):1--65, 1997.

\bibitem{icaps17}
Thomas~L. McCluskey and Mauro Vallati.
\newblock Embedding automated planning within urban traffic management
  operations.
\newblock In {\em Proceedings of the International Conference on Automated
  Planning and Scheduling {ICAPS}}, 2017.

\bibitem{minton}
Steven Minton.
\newblock Quantitative results concerning the utility of explanation-based
  learning.
\newblock In {\em AAAI}, pages 564--569, 1988.

\bibitem{ma}
M.~A.~H. Newton, J.~Levine, M.~Fox, and D.~Long.
\newblock Learning macro-actions for arbitrary planners and domains.
\newblock In {\em the International Conference on Automated Planning and
  Scheduling, ICAPS}, pages 256--263, 2007.

\bibitem{DBLP:journals/eswa/ParkinsonL15}
Simon Parkinson and Andrew~Peter Longstaff.
\newblock Multi-objective optimisation of machine tool error mapping using
  automated planning.
\newblock {\em Expert Syst. Appl.}, 42(6):3005--3015, 2015.

\bibitem{RiddleHB11}
Patricia~J. Riddle, Robert~C. Holte, and Michael~W. Barley.
\newblock Does representation matter in the planning competition?
\newblock In {\em Proceedings of the Ninth Symposium on Abstraction,
  Reformulation, and Approximation, {SARA} 2011, Parador de Cardona, Cardona,
  Catalonia, Spain, July 17-18, 2011.}, 2011.

\bibitem{madagascar}
J.~Rintanen.
\newblock Madagascar: Scalable planning with {SAT}.
\newblock In {\em Proceedings of the 8th International Planning Competition
  (IPC-2014)}, 2014.

\bibitem{use}
Reza Sadraei and Atefeh Ahmadi.
\newblock Use: The useful operator selection.
\newblock In {\em Proceedings of the 8th International Planning Competition
  (IPC-2014)}, 2014.

\bibitem{DBLP:conf/aaai/SeippSHH15}
Jendrik Seipp, Silvan Sievers, Malte Helmert, and Frank Hutter.
\newblock Automatic configuration of sequential planning portfolios.
\newblock In {\em Proceedings of the Twenty-Ninth {AAAI} Conference on
  Artificial Intelligence, January 25-30, 2015, Austin, Texas, {USA.}}, pages
  3364--3370, 2015.

\bibitem{thompson2012simple}
Steven~K Thompson.
\newblock Simple random sampling.
\newblock {\em Sampling, Third Edition}, pages 9--37, 2012.

\bibitem{arvandherd}
R.~Valenzano, H.~Nakhost, M.~M\"uller, and J.~Schaeffer.
\newblock Arvandherd 2014.
\newblock In {\em Proceedings of the 8th International Planning Competition
  (IPC-2014)}, 2014.

\bibitem{valenzano2014comparison}
Richard Valenzano, J~Schaeffer, N~Sturtevant, and Fan Xie.
\newblock A comparison of knowledge-based {GBFS} enhancements and
  knowledge-free exploration.
\newblock In {\em Proceedings of the 24th International Conference on Automated
  Planning and Scheduling (ICAPS-2014)}, pages 375--379, 2014.

\bibitem{VallatiFGHS13}
M.~Vallati, C.~Fawcett, A.~E. Gerevini, H.~H. Hoos, and A.~Saetti.
\newblock Automatic generation of efficient domain-optimized planners from
  generic parametrized planners.
\newblock In {\em Proceedings of the Sixth Annual Symposium on Combinatorial
  Search, {SOCS}}, 2013.

\bibitem{DBLP:journals/aim/VallatiCGMRS15}
Mauro Vallati, Luk{\'{a}}s Chrpa, Marek Grzes, Thomas~Leo McCluskey, Mark
  Roberts, and Scott Sanner.
\newblock The 2014 international planning competition: Progress and trends.
\newblock {\em {AI} Magazine}, 36(3):90--98, 2015.

\bibitem{DBLP:conf/socs/VallatiCM17}
Mauro Vallati, Luk{\'{a}}s Chrpa, and Thomas~Leo McCluskey.
\newblock Improving a planner's performance through online heuristic
  configuration of domain models.
\newblock In {\em Proceedings of the Tenth International Symposium on
  Combinatorial Search}, pages 171--172, 2017.

\bibitem{DBLP:conf/ijcai/VallatiHCM15}
Mauro Vallati, Frank Hutter, Luk{\'{a}}s Chrpa, and Thomas~Leo McCluskey.
\newblock On the effective configuration of planning domain models.
\newblock In {\em Proceedings of the Twenty-Fourth International Joint
  Conference on Artificial Intelligence, {IJCAI}}, pages 1704--1711, 2015.

\bibitem{yahsp3}
V.~Vidal.
\newblock Yahsp3 and yahsp3-mt in the 8th international planning competition.
\newblock In {\em Proceedings of the 8th International Planning Competition
  (IPC-2014)}, 2014.

\bibitem{wilcoxon}
F.~Wilcoxon.
\newblock Individual comparisons by ranking methods.
\newblock {\em Biometrics Bulletin}, 1(6):80--83, 1945.

\bibitem{jasper}
F.~Xie, M.~M\"uller, and R.~Holte.
\newblock Jasper: the art of exploration in greedy best first search.
\newblock In {\em Proceedings of the 8th International Planning Competition
  (IPC-2014)}, 2014.

\bibitem{YuanSB10}
Z.~Yuan, T.~St{\"{u}}tzle, and M.~Birattari.
\newblock Mads/f-race: Mesh adaptive direct search meets f-race.
\newblock In {\em Proceedings of the 23rd International Conference on
  Industrial Engineering and Other Applications of Applied Intelligent Systems
  ({IEA/AIE})}, pages 41--50, 2010.

\end{thebibliography}

\appendix
\section{Domain-by-domain results}\label{appendixdomain}

\begin{table}
\centering
\footnotesize
\begin{tabular}{l | cccc|cccc}
&   \multicolumn{4}{|c}{PAR10} &  \multicolumn{4}{|c}{\# Solved} \\
&  B & W & M & std  &  B & W & M & std \\
\hline
 \multicolumn{9}{c}{\bf Barman} \\
\hline
arvandherd  & 234.51 & 1057.22 & 639.48   & 189.53   &  20 &  14 &  17 & 2.38   \\
Bfs-f  & 9.18 & 174.98 & 18.56 & 55.69 & 20 & 19 & 20 & 0.20 \\
Cedalion  & 17.67 & 19.52 & 18.66   & 0.44   &  20 &  20 &  20 & 0.00   \\
Freelunch  & 3000.00 & 3000.00 & 3000.00   & 0.0   &  0 &  0 &  0 & 0.00   \\
IbaCoP  & 2443.71 & 2856.49 & 2854.68   & 78.28   &  4 &  1 &  1 & 0.38   \\
Jasper  & 28.94 & 486.71 & 183.76   & 129.38   &  20 &  17 &  19 & 1.21   \\
Mercury  & 1697.74 & 1970.14 & 1833.00  & 121.20   &  9 &  7 &  8 & 1.00   \\
Mpc  & 1846.09 & 2864.13 & 2573.74   & 237.65   &  8 &  1 &  3 & 2.26   \\
Probe  & 18.29 & 469.27 & 182.04   & 156.64   &  20 &  17 &  19 & 1.24   \\
SIW  & 3000.00 & 3000.00 & 3000.00   & 0.0   &  0 &  0 &  0 & 0.00   \\
use  & 20.44 & 23.30 & 21.73   & 2.23   &  20 &  20 &  20 & 0.00   \\
Yahsp3  & 2850.53 & 3000.00 & 3000.00   &  51.7  &  1 &  0 &  0 & 0.10   \\

\hline
 \multicolumn{9}{c}{\bf CaveDiving} \\
\hline
arvandherd  & 2011.25 & 2296.48 & 2017.93   & 80.30   &  7 &  5 &  7 & 0.23   \\
Bfs-f & 1950.42 & 2100.32 & 1955.41 & 37.41 &  7 &  6 & 6 & 0.12 \\
Cedalion  & 1975.99 & 1976.28 & 1976.14   & 0.06   &  7 &  7 &  7 & 0.00   \\
Freelunch  & 3000.00 & 3000.00 & 3000.00   & 0.0   &  0 &  0 &  0 & 0.00   \\
IbaCoP  & 1960.61 & 2128.32 & 1969.34   & 59.76   &  7 &  6 &  7 & 0.15   \\
Jasper  & 1952.47 & 1955.04 & 1953.04   & 1.34   &  7 &  7 &  7 & 0.00   \\
Mercury  & 2710.42 & 2711.04 & 2710.80  & 0.14   &  2 &  2 &  2 & 0.00   \\
Mpc  & 2121.25 & 3000.00 & 2712.42   & 204.62   &  6 &  0 &  2 & 1.41   \\
Probe  & 2250.67 & 3000.00 & 2700.62   & 303.57   &  5 &  0 &  2 & 2.00   \\
SIW  & 3000.00 & 3000.00 & 3000.00   & 0.0   &  0 &  0 &  0 & 0.00   \\
use  & 3000.00 & 3000.00 & 3000.00   & 0.0   &  0 &  0 &  0 & 0.00   \\
Yahsp3  & 3000.00 & 3000.00 & 3000.00   &  0.0  &  0 &  0 &  0 &  0.00  \\

\hline
 \multicolumn{9}{c}{\bf ChildSnack} \\
\hline
arvandherd  & 1846.03 & 2293.87 & 2016.06   & 142.01   &  8 &  5 &  7 & 1.12   \\
Bfs-f  & 1533.13 & 1806.29  & 1672.14 & 81.27 &  10 &  8 & 9 & 1.19 \\
Cedalion  & 3000.00 & 3000.00 & 3000.00   & 0.0   &  0 &  0 &  0 & 0.00   \\
Freelunch  & 2437.09 & 2573.62 & 2438.43   & 40.70   &  4 &  3 &  4 & 0.12   \\
IbaCoP  & 320.59 & 1206.87 & 763.76   & 181.01   &  18 &  12 &  15 & 1.41   \\
Jasper  & 3000.00 & 3000.00 & 3000.00   & 0.0   &  0 &  0 &  0 & 0.00  \\
Mercury  & 2283.96 & 2285.23 & 2284.33   & 0.31   &  5 &  5 &  5 & 0.00   \\
Mpc  & 1800.40 & 1953.30 & 1950.23   & 53.95   &  8 &  7 &  7 & 0.20   \\
Probe  & 3000.00 & 3000.00 & 3000.00   & 0.0   &  0 &  0 &  0 & 0.00   \\
SIW  & 3000.00 & 3000.00 & 3000.00   & 0.0   &  0 &  0 &  0 & 0.00   \\
use  & 1394.51 & 2115.26 & 1825.13   & 153.82   &  11 &  6 &  8 & 2.21   \\
Yahsp3  & 2550.05 & 3000.00 & 2850.03   &  124.94  &  3 &  0 &  1 & 1.08   \\
\hline
\end{tabular}

\caption{Results show the best (B), worst (W), median (M) PAR10 and coverage performance achieved on the IPC 2014 benchmarks, running planners on $50$ randomly configured domain models. ``std'' column reports the standard deviations. (PART 1)}
\end{table}

\begin{table}
\centering
\footnotesize
\begin{tabular}{l | cccc|cccc}
&   \multicolumn{4}{|c}{PAR10} &  \multicolumn{4}{|c}{\# Solved} \\
&  B & W & M & std  &  B & W & M & std \\
\hline
 \multicolumn{9}{c}{\bf CityCar} \\
\hline
arvandherd  & 203.09 & 788.74 & 498.80   & 161.13  &  19 &  15 &  17 & 1.41  \\
Bfs & 2116.23 & 2258.57 & 2253.84 & 41.08 &  6 &  5 & 5 & 0.18\\
Cedalion  & 2137.34 & 2138.13 & 2137.44   & 0.27   &  6 &  6 &  6 & 0.0   \\
Freelunch  & 3000.00 & 3000.00 & 3000.00   & 0.0   &  0 &  0 &  0 & 0.0   \\
IbaCoP  & 1713.18 & 2570.38 & 1997.55   & 223.69  &  9 &  3 &  7 & 2.41   \\
Jasper  & 2258.35 & 3000.00 & 2265.04   & 342.01  &  5 &  0 &  5 & 2.54   \\
Mercury  & 2430.31 & 2702.90 & 2429.63   & 77.40   &  4 &  2 &  4 & 0.15   \\
Mpc  & 1384.71 & 1811.22 & 1532.23   & 112.60  &  11 &  8 &  10 & 1.07   \\
Probe  & 763.78 & 2401.24 & 1662.58   & 373.69   &  15 &  4 &  9 & 2.45   \\
SIW  & 2255.80 & 2255.95 & 2255.87   & 0.04   &  5 &  5 &  5 & 0.00   \\
use  & 2721.95 & 3000.00 & 3000.00   & 73.78   &  2 &  0 &  0 & 0.20   \\
Yahsp3  & 3000.00 & 3000.00 & 3000.00 & 0.0 &  0 &  0 & 0 &  0.00    \\
\hline
 \multicolumn{9}{c}{\bf Floor} \\
\hline
arvandherd  & 2711.92 &  2859.32 & 2719.32  & 25.30   &  2 &   1 & 2& 1.00   \\
Bfs  & 2559.46 &  2850.15 & 2700.17  & 80.30   &  3 &  1 & 2 &  0.39   \\
Cedalion  & 2704.60 &  2704.74 & 2704.68  & 0.02   &  2 & 2 &  2 & 0.00   \\
Freelunch  & 2706.84 & 2707.21 & 2707.02  & 0.10   &  2 & 2 &  2 & 0.00   \\
IbaCoP  & 62.40 & 1088.96  & 63.14 & 235.57   &  20 &  13 & 20 & 2.43   \\
Jasper  & 2700.64 & 2702.23 & 2701.12  & 0.54   &  2 & 2 &  2 & 0.00   \\
Mercury  & 2700.75 & 3000.00 & 2700.78  & 113.80   &  2 &  0 & 2 & 0.00   \\
Mpc  & 0.11 &  0.12 & 0.11  & 0.00   &  20 & 20 &  20 & 0.00   \\
Probe  & 2709.62 &  2850.03 & 2800.12  & 47.71   &  2 & 1 &  1 & 0.14   \\
SIW  & 3000.00 & 3000.00 & 3000.00 & 0.00   &  0 & 0 &  0 & 0.00   \\
use  & 2579.21 &  3000.00  & 3000.0 & 138.49   &  3 & 0 &  0 & 1.00   \\
Yahsp3  & 2850.01  & 3000.00 & 3000.00  & 51.11   &  1 & 0 &  0 & 0.10   \\

\hline
 \multicolumn{9}{c}{\bf GED} \\
\hline
arvandherd  & 300.03 & 1200.03 & 900.03   & 253.11   &  18 &  12 &  14 & 1.73   \\
Bfs-f  & 1050.02 & 1800.04 & 1050.10   & 221.39   &  13 &  8 &  13 & 1.41   \\
Cedalion  & 0.04 & 0.06 & 0.05   & 0.02   &  20 &  20 &  20 & 0.00   \\
Freelunch  & 3000.00 & 3000.00 & 3000.00   & 0.00   &  0 &  0 &  0 & 0.00   \\
IbaCoP  & 3000.00 & 3000.00 & 3000.00   & 0.00   &  0 &  0 &  0 & 0.00   \\
Jasper  & 0.04 & 150.06 & 0.05   & 56.08   &  20 &  19 &  20 & 0.00   \\
Mercury  & 0.03 & 0.05 & 0.04   & 0.00   &  20 &  20 &  20 & 0.00   \\
Mpc  & 750.04 & 1350.04 & 1200.03   & 142.21   &  15 &  11 &  12 & 0.00   \\
Probe  & 3000.00 & 3000.00 & 3000.00   & 0.00   &  0 &  0 &  0 & 0.00   \\
SIW  & 2700.00 & 2850.00 & 2850.00   & 76.20   &  2 &  1 &  1 & 0.00   \\
use  & 0.03 & 0.05 & 0.04   & 0.00   &  20 &  20 &  20 & 0.00   \\
Yahsp3  & 1350.03 & 3000.00 & 3000.00   & 626.30   &  11 &  0 &  0 & 4.12   \\

\hline
\end{tabular}

\caption{Results show the best (B), worst (W), median (M) PAR10 and coverage performance achieved on the IPC 2014 benchmarks, running planners on $50$ randomly configured domain models. ``std'' column reports the standard deviations. (PART 2)}
\end{table}

\begin{table}
\centering
\footnotesize
\begin{tabular}{l | cccc|cccc}
&   \multicolumn{4}{|c}{PAR10} &  \multicolumn{4}{|c}{\# Solved} \\
&  B & W & M & std  &  B & W & M & std \\

\hline
 \multicolumn{9}{c}{\bf Hiking} \\
\hline
arvandherd  & 508.08 & 648.05 & 508.36  & 69.14   &  17 & 16 &  16 & 0.15   \\
Bfs-f  & 2705.51 &  2706.75 & 2705.80 & 0.28   &  2 & 2 &  2 & 0.00   \\
Cedalion  & 1013.75 &  1149.33 & 1013.82  & 54.35   &  14 & 13 &  13 & 0.18   \\
Freelunch  & 2402.08 &  2402.21 &  2402.13 & 0.02   &  4 & 4 &  4 & 0.00   \\
IbaCoP  & 786.94 &  937.97 & 790.65  & 36.45   &  15 & 14 &  14 & 0.10  \\
Jasper  & 968.55 & 1821.04 & 1256.85  & 220.79   &  14 &  8 & 12 & 2.41   \\
Mercury  & 1805.50 &  1805.63 & 1805.55  & 0.03   &  8 & 8  &  8 & 0.00   \\
Mpc  & 1956.28 &  2119.42 & 2103.43  & 81.45   &  7 & 6 &  6 & 0.69   \\
Probe  & 807.04 &  951.93 & 809.27  & 62.10   &  15 & 14 &  14 & 0.00   \\
SIW  & 2407.61 & 2550.93 & 2407.69  & 43.01   &  4 &  3 &  3 & 0.12   \\
use  & 1384.89 &  2256.60  & 1674.84 & 257.92   &  11 & 5 &  9 & 1.73   \\
Yahsp3  & 351.03 &  909.85 & 776.37  & 119.58   &  18 &  14 & 15 & 1.59   \\

\hline
 \multicolumn{9}{c}{\bf Maintenance} \\
\hline
arvandherd  & 629.61 & 1090.19 & 921.38   & 144.76   &  16 &  13 &  14 & 1.85   \\
Bfs-f  & 1808.98 & 1809.00 & 1809.00 & 0.01 &  8 &  8 & 8 & 0.0\\
Cedalion  & 3000.00 & 3000.00 & 3000.00   & 0.00   &  0 &  0 &  0 & 0.00   \\
Freelunch  & 17.38 & 17.57 & 17.41   & 0.05   &  20 &  20 &  20 & 0.00   \\
IbaCoP  & 487.89 & 491.54 & 491.44   & 1.25   &  17 &  17 &  17 & 0.00   \\
Jasper  & 926.74 & 1678.29 & 1520.27   & 244.17   &  14 &  9 &  10 & 1.41   \\
Mercury  & 1503.13 & 1503.42 & 1503.18   & 0.07   &  10 &  10 &  10 & 0.00   \\
Mpc  & 1055.52 & 1055.52 & 1055.70   & 0.18   &  13 &  13 &  13 & 0.00   \\
Probe  & 1950.38 & 1950.39 & 1950.39   & 0.01   &  7 &  7 &  7 & 0.00   \\
SIW  & 3000.00 & 3000.00 & 3000.00   & 0.00   &  0 &  0 &  0 & 0.00   \\
use  & 467.43 & 487.61 & 477.49   & 89.93   &  17 &  17 &  17 & 0.00   \\
Yahsp3  & 3000.00 & 3000.00 & 3000.00 & 0.0  &  0 &  0 &  0 & 0.00   \\

\hline
 \multicolumn{9}{c}{\bf Parking} \\
\hline
arvandherd  & 3000.00 & 3000.00 & 3000.00   & 0.00   &  0 &  0 &  0 & 0.00   \\
Bfs-f  & 2447.77 & 2721.20 & 2584.42   & 89.58   &  4 &  2 &  3 & 1.00   \\
Cedalion  & 2267.57 & 2267.75 & 2267.62   & 0.05   &  5 &  5 &  5 & 0.00   \\
Freelunch  & 3000.00 & 3000.00 & 3000.00   & 0.00   &  0 &  0 &  0 & 0.00   \\
IbaCoP  & 3000.00 & 3000.00 & 3000.00   & 0.00   &  0 &  0 &  0 & 0.00   \\
Jasper  & 1171.14 & 2164.24 & 1601.05   & 226.12   &  13 &  6 &  10 & 2.73   \\
Mercury  & 2726.28 & 2727.19 & 2726.71   & 0.24   &  2 &  2 &  2 & 0.00   \\
Mpc  & 2022.34 & 2440.27 & 2297.27   & 149.78   &  7 &  4 &  5 & 1.00   \\
Probe  & 2448.75 & 3000.00 & 2856.12   & 123.23   &  4 &  0 &  1 & 1.15   \\
SIW  & 657.37 & 952.03 & 807.00   & 81.61   &  16 &  14 &  15 & 0.40   \\
use  & 2714.02 & 3000.00 & 3000.00   & 61.53   &  2 &  0 &  0 & 0.20   \\
Yahsp3  & 3000.00 & 3000.00 & 3000.00   & 0.00   &  0 &  0 &  0 & 0.00   \\
\hline

\end{tabular}

\caption{Results show the best (B), worst (W), median (M) PAR10 and coverage performance achieved on the IPC 2014 benchmarks, running planners on $50$ randomly configured domain models. ``std'' column reports the standard deviations. (PART 3)}
\end{table}

\begin{table}
\centering
\footnotesize
\begin{tabular}{l | cccc|cccc}
&   \multicolumn{4}{|c}{PAR10} &  \multicolumn{4}{|c}{\# Solved} \\
&  B & W & M & std  &  B & W & M & std \\

\hline
 \multicolumn{9}{c}{\bf Tetris} \\
\hline
arvandherd  & 686.01 & 1252.67 & 1109.37   & 152.51   &  16 &  12 &  13 & 1.36   \\
Bfs-f  & 2705.81 & 2857.03 & 2707.85   & 37.91   &  2 &  1 &  2 & 0.10   \\
Cedalion  & 1589.46 & 1590.33 & 1589.81   & 0.30   &  10 &  10 &  10 & 0.00   \\
Freelunch  & 27.47 & 29.88 & 28.78   & 0.56   &  20 &  20 &  20 & 0.00   \\
IbaCoP  & 3000.00 & 3000.00 & 3000.00   & 0.00   &  0 &  0 &  0 & 0.00   \\
Jasper  & 1873.96 & 2704.57 & 2297.53   & 210.88   &  8 &  2 &  5 & 2.41   \\
Mercury  & 2427.87 & 2428.61 & 2428.39   & 0.21   &  4 &  4 &  4 & 0.00   \\
Mpc  & 1869.50 & 2296.57 & 2150.33   & 104.42   &  8 &  5 &  6 & 0.30   \\
Probe  & 1716.48 & 2862.10 & 2143.04   & 263.18   &  9 &  1 &  6 & 1.73   \\
SIW  & 2850.39 & 2854.22 & 2850.44   & 0.91   &  1 &  1 &  1 & 0.00   \\
use  & 2850.42 & 3000.00 & 3000.00   & 45.42   &  1 &  0 &  0 & 0.05   \\
Yahsp3 & 2850.48 & 2851.06 & 2850.52   & 0.10   &  1 &  1 &  1 & 0.00   \\

\hline
 \multicolumn{9}{c}{\bf Thoughtful} \\
\hline
arvandherd  & 344.53 & 392.53 & 364.50   & 81.67   &  18 &  18 &  18 & 0.00   \\
Bfs-f  & 614.78 & 614.80 & 614.95   & 0.20   &  16 &  16 &  16 & 0.00   \\
Cedalion  & 32.52 & 32.46 & 32.59   & 0.06   &  20 &  20 &  20 & 0.00   \\
Freelunch  & 2864.15 & 2864.29 & 2864.20   & 0.03   &  1 &  1 &  1 & 0.00   \\
IbaCoP  & 3000.00 & 3000.00 & 3000.00   & 0.00   &  0 &  0 &  0 & 0.00   \\
Jasper  & 458.29 & 1055.73 & 758.15   & 169.76   &  17 &  13 &  15 & 1.59   \\
Mercury  & 1207.07 & 1207.31 & 1207.18   & 0.05   &  12 &  12 &  12 & 0.00   \\
Mpc  & 2250.52 & 2250.64 & 2250.57   & 0.03   &  5 &  5 &  5 & 0.00   \\
Probe  & 474.80 & 1217.47 & 920.53   & 182.20   &  17 &  12 &  14 & 2.21   \\
SIW  & 152.96 & 301.51 & 301.50   & 71.36   &  19 &  18 &  18 & 0.00   \\
use  & 2850.06 & 3000.00 & 3000.00   & 37.45   &  1 &  0 &  0 & 0.08   \\
Yahsp3  & 20.92 & 171.63 & 22.97   & 36.95   &  20 &  19 &  20 & 0.09   \\

\hline
 \multicolumn{9}{c}{\bf Transport} \\
\hline
arvandherd  & 2125.34 & 2422.90 & 2270.05   & 72.74   &  6 &  4 &  5 & 0.20   \\
Bfs-f  & 2575.35 & 2576.05 & 2575.57   & 0.16   &  3 &  3 &  3 & 0.00   \\
Cedalion  & 87.58 & 88.67 & 87.93   & 0.34   &  20 &  20 &  20 & 0.00   \\
Freelunch  & 3000.00 & 3000.00 & 3000.00   & 0.00   &  0 &  0 &  0 & 0.00   \\
IbaCoP  & 3000.00 & 3000.00 & 3000.00   & 0.00   &  0 &  0 &  0 & 0.00   \\
Jasper  & 2286.67 & 2422.72 & 2422.09   & 67.92   &  5 &  4 &  4 & 0.10   \\
Mercury  & 47.91 & 49.64 & 48.69   & 0.44   &  20 &  20 &  20 & 0.00   \\
Mpc  & 3000.00 & 3000.00 & 3000.00   & 0.00   &  0 &  0 &  0 & 0.00   \\
Probe  & 2146.58 & 2289.36 & 2152.69   & 69.01   &  6 &  5 &  6 & 0.10   \\
SIW  & 404.02 & 696.94 & 405.88   & 135.17   &  18 &  16 &  18 & 0.05   \\
use  & 3000.00 & 3000.00 & 3000.00   & 0.00   &  0 &  0 &  0 & 0.00   \\
Yahsp3  & 11.01 & 14.81 & 15.14   & 1.29   &  20 &  20 &  20 & 0.00   \\

\end{tabular}

\caption{Results show the best (B), worst (W), median (M) PAR10 and coverage performance achieved on the IPC 2014 benchmarks, running planners on $50$ randomly configured domain models. ``std'' column reports the standard deviations. (PART 4)}
\end{table}

\begin{table}
\centering
\footnotesize
\begin{tabular}{l | cccc|cccc}
&   \multicolumn{4}{|c}{PAR10} &  \multicolumn{4}{|c}{\# Solved} \\
&  B & W & M & std  &  B & W & M & std \\

\hline
 \multicolumn{9}{c}{\bf Visitall} \\
\hline
arvandherd  & 2434.78 & 2435.65 & 2435.37   & 0.22   &  4 &  4 &  4 & 0.00   \\
Bfs-f  & 2266.40 & 2266.98 & 2266.64   & 0.22   &  5 &  5 &  5 & 0.00   \\
Cedalion  & 6.79 & 7.02 & 6.88   & 0.06   &  20 &  20 &  20 & 0.00   \\
Freelunch  & 2101.71 & 2101.84 & 2101.72   & 0.03   &  6 &  6 &  6 & 0.00   \\
IbaCoP  & 3000.00 & 3000.00 & 3000.00   & 0.00   &  0 &  0 &  0 & 0.00   \\
Jasper  & 1860.37 & 2271.11 & 2132.25   & 146.50   &  8 &  5 &  6 & 1.35   \\
Mercury  & 1712.02 & 1712.55 & 1712.27   & 0.37   &  9 &  9 &  9 & 0.00   \\
Mpc  & 3000.00 & 3000.00 & 3000.00   & 0.00   &  0 &  0 &  0 & 0.00   \\
Probe  & 2427.55 & 2427.98 & 2427.74   & 0.14   &  4 &  4 &  4 & 0.00   \\
SIW  & 1141.64 & 1142.25 & 1141.65   & 0.15   &  13 &  13 &  13 & 0.00   \\
use  & 3000.00 & 3000.00 & 3000.00   & 0.00   &  0 &  0 &  0 & 0.00   \\
Yahsp3  & 1701.89 & 1702.18 & 1702.12   & 0.18   &  9 &  9 &  9 & 0.00   \\

\hline

\end{tabular}

\caption{Results show the best (B), worst (W), median (M) PAR10 and coverage performance achieved on the IPC 2014 benchmarks, running planners on $50$ randomly configured domain models. ``std'' column reports the standard deviations. (PART 5)}
\end{table}


\end{document}